\def\gmode{} %%use this to include figures

 \def\gdriver{dviout}%for dviout for preview

\newcommand{\dtitle}[1]{\title{ \if \gmode \else
\color{red} Demo mode!\\
comment out \textbackslash def \textbackslash gmode\{demo\} at the header to include figures \color{black}\\
\fi
#1 }}
\ifx\pdfoutput\undefined
\else
 \def\gdriver{}
\fi
\documentclass[\gmode,\gdriver,10pt,twocolumn,letterpaper]{article}

\usepackage{3dv}
\usepackage{times}
\usepackage{epsfig}
\usepackage{graphicx}
\usepackage{amsmath}
\usepackage{amssymb}
\usepackage{color}
\usepackage{here}

\threedvfinalcopy % *** Uncomment this line for the final submission

\def\threedvPaperID{206} % *** Enter the 3DV Paper ID here

\ifthreedvfinal\fi
\usepackage{comment}
\newcommand{\fnote}[1]{}

\newcommand{\knote}[1]{{\color{red} \bf #1 \color{black}}}

\newcommand{\bnote}[1]{{\color{red} #1 \color{black}}}
\newcommand{\mnote}[1]{}
\newcommand{\onote}[1]{\color{blue} #1 \color{black}}
\newcommand{\snote}[1]{}
\newcommand{\kcut}[1]{}
\newcommand{\ocut}[1]{}
\newcommand{\scut}[1]{}
\newcommand{\jptext}[1]{}

\ifx\pdfoutput\undefined
\else
 \usepackage{CJKutf8}
 \renewcommand{\fnote}[1]{}
 \renewcommand{\knote}[1]{}
 \renewcommand{\mnote}[1]{}
 \renewcommand{\bnote}[1]{}
 \renewcommand{\onote}[1]{}
 \renewcommand{\jptext}[1]{}
\fi

\ifx\etal\undefined\newcommand{\etal}{{\it et al. }}\fi
\ifx\ie\undefined\\newcommand{\ie}{{\it i.e.}}\fi
\ifx\eg\undefined\\newcommand{\eg}{{\it e.g.}}\fi

\newcommand{\subsecref}[1]{Sec.~\ref{ssec:#1}}

\graphicspath{{./figure/}{../181007-icip-underwater-active-stereo/}{../180908-eccv-underwater-cnn-active-stereo/}}

\begin{document}

%%%%%%%%% TITLE
\dtitle{Multi-scale CNN stereo and pattern removal technique \\for underwater active stereo system} % Replace with your title

\author{Kazuto Ichimaru$^{\dag}$ \,\,\, Ryo Furukawa$^{\ddag}$ \,\,\, Hiroshi Kawasaki$^{\dag}$\\
$^{\dag}$ Kyushu University, Fukuoka, Japan\\
$^{\ddag}$ Hiroshima City University, Hiroshima, Japan}

\maketitle\thispagestyle{empty}

\begin{abstract}
Demands on capturing dynamic scenes of underwater environments are rapidly growing.
Passive stereo is applicable to capture dynamic scenes, however 
%it is not     commonly used for practical purposes, because 
the shape with textureless 
    surfaces or irregular reflections cannot be recovered by the technique.
In our system, we add a pattern projector 
    to the stereo camera pair so that artificial textures are augmented on the 
    objects.
To use the system at underwater environments, 
several problems should be compensated, \ie, refraction, disturbance 
    by fluctuation and bubbles. Further, since surface of the objects are interfered by the bubbles, projected 
    patterns, etc., those noises and patterns should be removed from captured 
    images to recover original texture. % with pattern illumination. %our system.
To solve these problems, we propose three approaches;
a depth-dependent calibration, 
%a CNN-based target object segmentation method, CNN-based stereo method and 
Convolutional Neural Network(CNN)-stereo method and 
    CNN-based texture recovery method.
A depth-dependent calibration is our analysis to find the acceptable depth range 
    for approximation by center projection to find the certain target depth for calibration.
% based on our analysis that  approximation 
%    by center projection can be applied to a wide depth range.
% for practical applications.
In terms of CNN stereo, unlike common CNN-based stereo methods which do not consider strong 
    disturbances like refraction or bubbles, we designed a novel CNN architecture 
    for stereo matching using multi-scale information, which is intended 
to be robust against such disturbances. %underwater disturbances.
Finally, we propose a multi-scale method for bubble and a projected-pattern removal method
%and %image based a bubble-canceling method 
using CNNs to recover original textures.
Experimental results are shown to prove the effectiveness of our method compared 
    with the state of the art techniques.
Furthermore, reconstruction of a live swimming fish is demonstrated to confirm the feasibility %and practicability
 of our techniques.
%
%\begin{keywords}
%Calibration, Object segmentation, Stereo matching, 3D reconstruction
%\end{keywords}
%

\end{abstract}

\section{Introduction}
\label{sec:intro}

There are strong demands on capturing dynamic scenes of underwater environments, 
\eg, measurement of seabeds, capturing dynamic shape deformations of swimming 
fish or humans, 
inspection of water-filled nuclear tanks by autonomous robots, etc.
Passive stereo is a common solution for capturing 3D shapes because 
of its great advantage of simplicity; \ie, it only requires two cameras in 
theory. %for capturing.
% which are captured at the same time.
In addition, since the shapes are recovered only from a pair of stereo images, it can 
capture moving or deforming objects.
%, this is another important advantage of the technique.
%also . The capability to capture such objects is called ``one-shot."
One severe problem on passive stereo is instability, \ie, it fails to capture 
objects with textureless surfaces or irregular reflection.
To overcome the problem, using a pattern projector to add an artificial texture 
onto the objects has been proposed~\cite{konolige}. In the system, we also take the same approach to 
achieve robust and dense reconstruction.
% in our technique.
%for our system. 

Considering underwater environments, there are additional problems for shape 
reconstruction by stereo, such as %light attenuation, 
refraction and disturbances by fluctuation and bubbles.
Further, since original textures of objects are interfered by projected 
    patterns if active illumination is projected, they should be removed for obtaining
    both 3D shapes and textures. %simultaneously.
In this paper, we propose three approaches to solve aforementioned problems.
For the refraction issue,  a depth-dependent calibration where refractions are 
approximated by lens distortion of a center projection model 
 is proposed~\cite{Kawasaki:WACV17}.
In the paper, we analyze to find the acceptable depth range 
    for the approximation and find the best depth for calibration.
For the problems of disturbances by obstacles, we propose Convolutional Neural Network(CNN)-based stereo
%approaches 
as a solution.
Since captured images of underwater scenes are affected by 
mixtures of
light attenuation caused by strong absorption of light intensity in water 
medium and strong
disturbances such as %, which are caused by %mainly three reasons, such as 
bubbles, shadows of water surface or fluctuation,
it is impossible to decompose them analytically. %analyze all of them. %
%light attenuation is caused by strong absorption of light intensity by water 
%medium and 
%disturbances are caused by %mainly three reasons, such as 
%bubbles, shadows of water surface or fluctuation of water volume,
%final representation of underwater scene is a complicatedly 
%mixtured of those phenomena and impossible to be decomposed analytically.
% solution is difficult to be applied. 
%
To handle such difficult problems, learning-based approaches, especially 
CNN techniques, are proposed.
% and a number of promising results were shown.
% draw a wide attention. 
%and
%In this paper, 
%we also propose a CNN-based approach. The technique consists
% of two 
%parts, 

Our shape reconstruction consists of two techniques, such as
%the one is based on 
%target object segmentation by CNN, and the other is patch-based CNN stereo.
CNN-based object segmentation and 
CNN-based stereo matching.
%and CNN-based texture recovery.
%Note that those techniques independently work efficiently, however, 
%works better if together.
%
The CNN-based target object segmentation method efficiently segment a target object, 
\eg, fish in our experiment, from background, which is not only useful for reducing calculation times, 
but also effective to achieve robust reconstruction by 
narrowing the search ranges of stereo disparities.
CNN-based stereo effectively works under common variations~\cite{mccnn}, however, 
there are strong disturbances at underwater environment.
%strong noise disturbances, where
%the architecture is inspired by \cite{mccnn}, the technique is designed to be 
%robust against noise.
In case of such strong disturbances, 
we propose a novel architecture of CNN, which uses multi-scale information of 
captured images.
\iffalse
The CNN-based object segmentation is based on U-Net~\cite{ronneberger2015u}
and.
CNN-based stereo is based on \cite{mccnn}, however, the technique is originally 
designed to be 
robust against noise by multi-scale extension.
%strong noise disturbances, where
%the architecture is inspired by \cite{mccnn}, the technique is designed to be 
%robust against noise.
\fi

%We also In case disturbances is too large obstacles like bubbles. For solution,
%we designed a novel architecture of CNN, which uses multi-scale information of 
%captured image.
%CNN-based stereo matching called Multi-scale CNN Stereo for the purpose.
% (e.g. bubbles or shadows of water surfaces).
For the texture recovery, we also propose a CNN-based method for projected-pattern removal and %image based 
    bubble cancellation.
% which consists of input and ground truth. of  trained by real and CG synthesized data set.
Main contributions of the proposed technique are as follows:
\begin{enumerate}
  \setlength{\parskip}{0cm} % 段落間
  \setlength{\itemsep}{0cm} % 項目間
\item A practical technique is proposed to achieve dense and robust shape reconstruction  
      based on passive stereo using active pattern projection.
\item A valid depth range for depth-dependent approximation by radial distortion is analysed.
\item A target-region detection method by CNN for robust stereo matching is proposed.
\item A multi-scale CNN-based stereo technique specialized for underwater environment is proposed.
\item A multi-scale CNN-based bubble and projected pattern removal method specialized for 
		  underwater environment is proposed.
\end{enumerate}
%\begin{itemize}
%	\item Accurate and dense underwater one-shot reconstruction with a passive active stereo approach.
%	\item Target region detection method by a CNN  for robust stereo matching.
%	\item A CNN-based stereo matching that is robust against underwater disturbances (e.g. bubbles or shadows of water surfaces).
%	\item Transfer learning of CNN-based stereo to improve its robustness against such disturbance.
%\end{itemize}
Experimental results are shown to prove the effectiveness of our method by 
comparing the results with the previous method. 
%ground truth. 
We also conduct demonstration to show the reconstructed sequence of a  swimming fish.

\section{Related works}
\label{sec:related}

%%\knoteII{Make this section more concise, maybe half or 3/4.}

To recover shape and texture of underwater environment, many researches have 
been done.
Main issue for underwater environment is refraction and generally two types of solution are proposed;
one is geometric approach and the other is approximation-based approach.
Geometric approach is based on physical models such as refractive index, distance to refraction interface, and normal of the interface.
%~\cite{Agrawal:CVPR2012,jordt2013refractive,kawahara:ICCV2013}.
Agrawal \etal introduced polynomial formulation for the model~\cite{Agrawal:CVPR2012}.
Sedlazeck and Koch proposed structure from motion for underwater environment~\cite{jordt2013refractive}.
Kawahara \etal proposed pixel-wise varifocal camera model~\cite{kawahara:ICCV2013}.
In this model, appropriate focal lengths are assigned to each pixel.
%They also proposed a reconstruction method with the model, but reconstruction result was not precise enough because it used space carving method with small number of cameras\cite{kawahara2}.
Those techniques can calculate genuine light rays if parameters are correctly estimated and 
interface is completely planar, however, they are usually impractical. 
%Further, 
%the non-central projection camera model is not suitable for shape 
%reconstruction in theory.
%
On the other hand, approximation approach converts captured images into central 
projection images by lens distortion and focal length adjustment~\cite{ferreira2005stereo}.
They assumed focal point moved backward to adjust light paths as linear as possible, then remaining error was treated as lens distortion.
%Because calculation cost of such method is very small compared to geometric approach, we used it in this work.
Kawasaki \etal. also proposed a simple method to approximate the refraction by radial 
distortion~\cite{Kawasaki:WACV17}. Since the parameter cannot be fixed for all 
the depth range, they proposed a depth dependent technique.
%Because calculation cost of such methods is small compared to geometric approach, 
%we used the model in our work.
It works well in most cases, however in specific case it fails because refractive 
distortion depends on depth and effective range of depth is not thoroughly
analyzed yet.
%However in fact, depth dependency has little effect on accuracy.
%It can be simply ignored and error ratio is still kept low.
%In this work, we used approximative approach and ignored depth dependency.

\ifx
There are two major problems in underwater capturing: refraction and attenuation.
For refraction problem, there are generally two types of solution;
one is geometric approach and the other is approximative approach.
Geometric approach considers physical models which represents light transmission.
%They have some unknown parameters such as refractive index, distance to refraction surface, and normal of the surface.
Unknown parameters, such as refractive index, distance to refraction surface, and normal of the surface, can be retrieved by calibration, but in general the more unknown parameters exist, the more calculation cost increases.
Agrawal \etal. introduced a physical model under flat refraction constraint and proposed a calibration method, but it needs to solve 4th degree equation even in one refractive surface situation, and 12th degree equation in two refractive surface situation\cite{Agrawal:CVPR2012}.
Sedlazeck and Koch considered pixel-wise virtual camera and proposed efficient error functions to reconstruct underwater object with Structure-from-Motion, but still nonlinear optimization with many parameters is required\cite{jordt2013refractive}.
Kawahara et al. proposed a non-central projection model called Pixel-wise Varifocal Camera Model to represent efficient forward projection\cite{kawahara:ICCV2013}.
%In this model, appropriate focal lengths are assigned to each pixel.
They also proposed a reconstruction method with the model, but reconstruction result was not precise enough because it used space carving method with small number of cameras\cite{Kawahara:CVA2014}.

On the other hand, approximative approach uses lens distortion and focal length adjustment as approximation to refractive distortion.
Ferreira et al. introduced them into passive stereo reconstruction\cite{ferreira}.
They assumed focal point moved backward to adjust light paths to be as linear as possible, then the remaining errors were treated as lens distortions.
Because calculation cost of such method is very small compared to geometric approach, we used it in this work.
%It works well in most cases, but in specific case it fails because refractive distortion depends on depth and it is not considered.
%However in fact, depth dependency has little effect on accuracy.
%It can be simply ignored and error ratio is still kept low.
%In this work, we used approximative approach and ignored depth dependency.
\fi

\ifx
In terms of light attenuation and disturbance problem for water medium,
light transport analysis has been conducted~\cite{Kutulakos:PAMI16,mukaigawa2010analysis}.
%For active lighting method such as structured light for underwater, usually simpler
%solution is adopted, \ie, high power laser is used.
Narasimhan \etal proposed a structured-light-based 3D scanning method for 
strong scattering and absorption media based on light transport analysis~\cite{narasimhan2005structured}.
For weak scattering media, Bleier and N\"uchter used cross laser projector which only achieved sparse reconstruction~\cite{bleier}.
To increase density, Campos and Codina projected parallel lines with DOE to capture underwater objects with one-shot scan~\cite{massot2014underwater}.
Kawasaki \etal. proposed a grid pattern to capture more dense 
shape with one-shot scan~\cite{Kawasaki:WACV17}.
% with DOE with special calibration techniques~\cite{Kawasaki:WACV17}. 
One drawback of those one-shot scanning techniques is that 
reconstruction tends to be unstable even if light attenuation and disturbances 
are not so strong because sensitivity of pattern detection is high for 
subtle change of projected pattern.
%getting stronger with stronger scattering media.
% because pattern detection is severely affected 
%by light attenuation and disturbances.
\fi

\ifx
However, attenuation problem still remains.
In the water, visible light attenuates rapidly as depth increases compared to in the air.
%Akkaynak et al. showed light intensity decreased to at most about 50\% at a distance of 20m\cite{atten}.
Then stereo matching becomes difficult because there is little contrast left.
To solve this problem, simply additional light source is effective.
In active stereo method, projected pattern works as additional light and it becomes easier to search correspondence, but lens distortion approximation of refracted pattern is difficult.
One solution to cope with this problem is to use pattern which are hardly affected by refraction, such as line laser or diffractive optical element(DOE).
Such patterns surely works in the water, but they often sacrifices one-shot or reconstruction density.
Bleier and N\"uchter used cross laser projector to avoid refraction effect, but the method is not one-shot~\cite{bleier}.
Campos and Codina projected line pattern with DOE to capture underwater object with one-shot active stereo~\cite{massot2014underwater}.
Kawasaki et al. also projected wave pattern with DOE to do the same thing, but both of the patterns was not dense enough to reconstruct dense mesh~\cite{Kawasaki:WACV17}.
\fi

\ifx
%Passive stereo is also a major solution for underwater scanning.
One simple solution is to apply passive stereo which is not much affected by those effects.
% is also a major solution for underwater scanning.
In the air, to increase the stability, Konolige investigated how to add active pattern to 
the passive stereo system~\cite{konolige}. 
Until now, there are no major passive stereo with active lighting system proposed for underwater environment yet.
%In the field of stereo matching, semi global block matching (SGBM) is one of the 
%most popular methods~\cite{SGBM}, which is proposed by 
%Hirschm\"uller, but not enough robust to noises.
\fi

\jptext{
CNNステレオが提案されている。例えば、Zbontarの手法はパッチの類似性を算出する[0]。
これは、テクスチャがないところで動くとされる。
パッチベースのCNNステレオは遅いので、その高速化手法が研究されている。
[2]は、最後のFC層を内積に置き換えることで、速度を早める方法を提案した。
[3]は、FC層と内積を組み合わせることで、精度を維持しながら、速度を早めることに成功している。
根本的な速度の改善のために、通称DispNetと呼ばれている、End-to-endでする手法も提案されている[1]。
しかし、精度は高くない。
そこで、我々はパッチベース手法を用いる。
パッチベースの手法はノイズには強いが構造が大きな障害物には弱いため、我々は、Multi-scaleに
してこの問題を解消した。Multi-scale cnnは他の分野では最近研究されるようになってきた。
でブラーでは[10]や、・・・・など。
我々も同様のアーキテクチャを採用する。
}

Another problem for underwater environment is disturbances by bubbles, water 
fluctuation and other effects.
Recently, convolutional neural network (CNN) based stereo matching becomes 
popular, which is robust to irregular distortion on image set.
\u{Z}bontar and LeCun proposed a CNN-based method to train network as a cost function of image patches~\cite{mccnn}.
Those techniques rather concentrate on textureless region recovery, but not 
noise compensation, which is a main problem for underwater stereo.
Since patch based technique is known to be slow, Luo \etal proposed a speeding-up technique by substituting FCN to inner product at final 
stage~\cite{Luo:CVPR2016}. Shaked and Wolf achieved high accuracy as well as fast 
calculation time by combining both FCN to inner product~\cite{Shaked:CVPR2017}.
To fundamentally solve the calculation time, end-to-end approach called DispNet is proposed, but accuracy is not so high~\cite{Mayer:CVPR2016}.
%Another problem for patch based CNN stereo is that it is severely affected by 
%obstacles, image degradation or various scaling.
% because enlarge the patch size is theoretically difficult. 
Another aspect for underwater environment is that range of the scale of 
obstacles is large.
Recently, to solve such scaling problem, multi-scale CNN 
technique is proposed. Nah \etal proposed a 
method for deblurring~\cite{Nah:CVPR2017}, Zhaowei \etal proposed a method for 
dehaze~\cite{Cai2016AUM} and Li \etal proposed a method for object recognition~\cite{li2017reside},
%however there is no multi-scale technique for stereo yet with the best of our knowledge.
Yadati \etal, Lu \etal, and Chen \etal~\cite{PramodYadati:ICMVA2017,HaihuaLu2018,JiahuiChen:ICIP2016} used multi-scale features for CNN-based stereo matching.
We also use multi-scale features for CNN-based stereo matching,
but novel network architecture to recognize multi-scale information is proposed.

\jptext{
また、CNNステレオは、学習データの入手が困難という問題もある。
[4]では、教師無しでEnd-to-Endネットワークを構築する。これは、ロス関数に
明示的な関数を用いないことで実現される（How?）。
[5]は、既存のステレオマッチングアルゴリズムを教師とする半教師あり学習である。
[6]は、様々な制約条件を用いたコスト関数を用いてMIL(Multi-Instance Learning)を行う。
我々と、精度と効率を考え、半教師あり学習となっている。
}

Collection of huge data for learning is another open problem for CNN-based stereo 
techniques. For solution, Zhou \etal proposed a technique without using ground 
truth depth data, but LR consistency as a loss function~\cite{Zhou:ICCV2017}. Tonioni \etal 
proposed a unsupervised method by using existing stereo technique as an 
instruction~\cite{Tonioni:ICCV2017}.
Tulyakov and Ivanov proposed a multi-instance learning (MIL) method by using 
several constraints and cost functions~\cite{Tulyakov:ICCV2017}. We also take a 
similar approach to \cite{Tulyakov:ICCV2017} and use several cost functions.

\ifx
Passive active stereo can solve such problems.
It projects a pattern, but the correspondence between feature points on pattern and captured images are neither used in calibration nor reconstruction. 
%Pattern only works as matching features, that is, image to image correspondence.
Therefore, refraction on projected pattern can be remained, and ``rich" pattern can be used.
For example, Konolige investigated how passive active stereo works\cite{konolige}.
They tried several types of patterns and showed optimized pattern works better than not optimized ones. 
For the above reasons, we took passive active stereo method.

In the field of stereo matching, Semi Global Block Matching(SGBM) is one of the most popular methods.
Hirschm\"uller proposed semi-global matching using mutual information\cite{SGBM}, but it was not robust enough against hard cases such as bubbles without post-processing like interpolation.
Recently, CNN-based stereo matching became popular.
\u{Z}bontar and LeCun proposed a CNN-based method to train network as a cost function of image patches\cite{mccnn}.
Although they did not mention robustness of CNN-based stereo against such disturbance, we consider it has ability to avoid it.
We show how effective CNN-based stereo is against disturbance, in our case, bubbles.
%In addition, we tried transfer-learning to improve its robustness against bubbles using dataset we made from existing stereo dataset.
\fi

\jptext{
提案手法では、パターンを投影するため、これを除去したり、これを使ったセグメンテーションを行う。
セグメンテーションはSegNetやUnetがポピュラーであり、我々もこれを使う。
パターン除去には、GANによるインペインティングが有効と思われるが[11]、生成的アプローチは解像度が低いので今回は
使用しない。そこで、パターンをノイズと考え除去する。
CNNによるノイズの除去としては、Deepなアーキテクチャではなく訓練が簡単なResidualを用いることで、
訓練しやすいアーキテクチャである、ResBlockが提案された[8]。
このResBlockを用いて、反射除去のためにCEILNetというアーキテクチャも提案されている[9]。
さらに、WIN5RBネットワークを用いてEnd-to-Endでより効率よくノイズ除去する手法も提案されている[7]。
本研究ではこの[7]を用いてあぶく除去を行っているが、学習データの取得方法にオリジナリティがある。
}
CNNs are also popular in the field of image restoration and segmentation.
%We show how segmentation works to clean reconstruction result.
In underwater environment, there are several noises, such as bubbles or shadows 
of water surfaces. In addition, projected pattern onto the target object is also 
a severe noise.
To remove such a large noise,  inpainting method based on a GAN is 
promising~\cite{Iizuka:SIGGRAPH16,ChenyuYou2018}. However, since resolution of generative approaches are 
basically low, noise removal approach is better fit to our purpose. For efficient 
noise removal, shallow CNN-based approach using residual is 
proposed~\cite{He:CVPR2016}. The technique is also extended to remove 
reflection~\cite{Fan:CVPR2017}. Liu and Fang propose an end-to-end architecture using the 
WIN5RB network~\cite{Peng:arXiv2017} which outperform others. We also use this 
technique, but data collection and multi-scale extension is novel. %contribution.
Liao \etal~\cite{XuanLiao:VCIP2017} denoised depth images both using depth image and RGB image. 
Zhang \etal~\cite{KaiZhang2017} denoised images with CNNs with different noise levels taken into account. 
Choi \etal~\cite{SungjoonChoi2018} proposed denoising with multi-scales with light-weight computation.
Nakamura \etal~\cite{ToshikiNakamura:ICDAR2017} removed texts in natural scene images using multi-layers of convolutions and
deconvolutions. 

Image segmentation is also important for our system, since usually only the regions of the target 
object are enough for 3D shape reconstruction. 
Badrinarayanan \etal. proposed a network architecture for semantic segmentation called SegNet~\cite{SegNet}.
%It is end-to-end network which takes natural scenery images as input and output color masks classified into multi class.
Ronneberger \etal also proposed a network architecture called U-Net which is useful for biomedical image segmentation~\cite{ronneberger2015u}.
Since captured images do not look similar to scenery image, but rather close 
to biomedical image, we use U-Net for our segmentation.

%\section{Proposal Method}
%\label{sec:proposal}
%In this section, we introduce our proposal method.
%First, we mention system configuration and algorithm.
%Second, implementation techniques are described.

\section{System and algorithm overview}
\label{sec:overview}
%Key points of our method is underwater passive active stereo and CNN-based stereo.
%First, we introduce overview of the method in this section, and describe implementation techniques in the following section.\ref{sec:cnnstereo}.

\subsection{System Configuration}
\label{ssec:sysconf}
\begin{figure}[t]

\begin{minipage}[b]{.49\linewidth}
  \centering
%  \centerline{\includegraphics[width=4.0cm]{images/sysconf2.eps}}
%  \centerline{\includegraphics[width=4.3cm]{images/systemconfig.eps}}
  \centerline{\includegraphics[width=3.5cm]{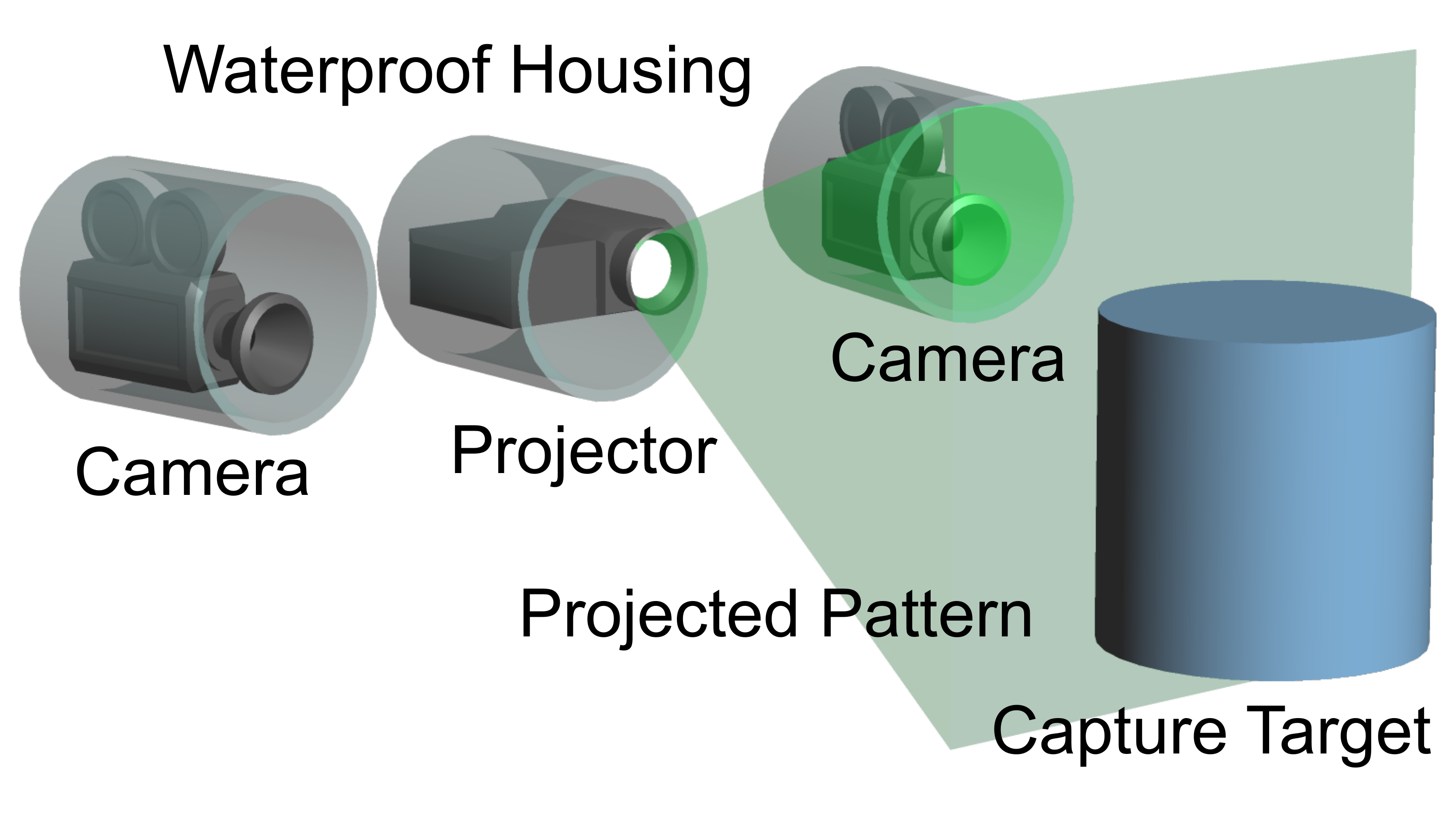}}
\end{minipage}
\hfill
\begin{minipage}[b]{0.49\linewidth}
  \centering
%  \centerline{\includegraphics[width=3.5cm]{images/experiment/jpg/IMG_0200.jpg}}
%  \centerline{\includegraphics[width=3.5cm]{images/real_system.eps}}
  \centerline{\includegraphics[width=4.3cm]{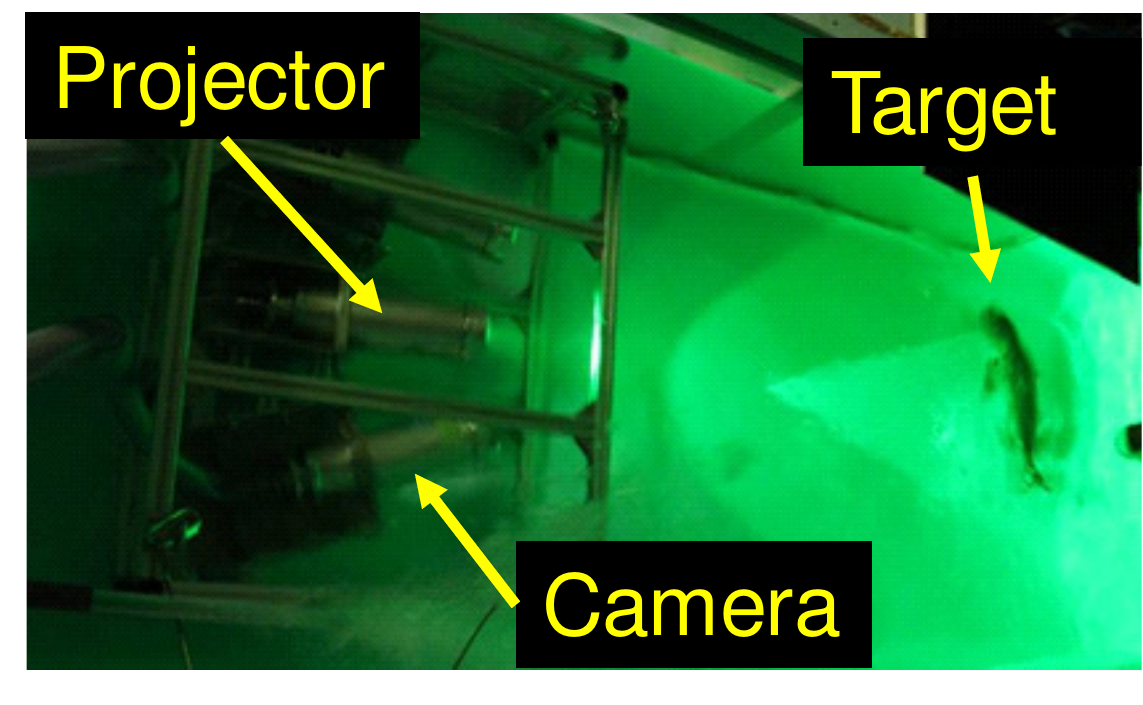}}
\end{minipage}

\caption{{\bf Left:} Minimum system configuration of the proposed algorithm. {\bf Right:} Our 
    experimental system for evaluation where two cameras and a projector are set outside a water tank.}
\label{fig:sysconf}
\end{figure}

Our system consists of
%Minimum system configuration is 
stereo camera pair and one laser projector as shown in Fig.~\ref{fig:sysconf}.
We prepare two systems for our experiments.
One is for evaluation purpose where two cameras and a projector are set outside a water tank.
The other is a practical system where devices are installed into a specially built waterproof 
housing in order to make distance between interface glass and camera lens to be 
relatively short. For the both systems, the optical axes of the cameras are set orthogonal to glass surface so 
that error by refraction approximation is minimized. 
%Cameras and projector are adjusted to have their field-of-views overlap each other(This is prerequisite of stereo method).
The two cameras are synchronized by GPIO cable to capture dynamic scenes.
In terms of the pattern projector to add textures onto the objects,
no synchronization is required
since the pattern is static. 
In our implementation, we use a laser projector where diffractive optic element 
(DOE) is used to configure wave pattern proposed in \cite{Sagawa:3DIMPVT2012} 
without losing light power. % for it.
%Since we use the pattern for only adding textures to the scene, 
%it can be any patterns and may be refracted. 
% 
% since correspondences between projected pattern and captured images are not used, the pattern is not restricted at all.
%The housing containing cameras and projector is submerged and fixed.

%For calibration, some kind of calibration tools, such as calibration board or calibration box, can be used. 
%We used waterproof calibration board with AR pattern.

%(system confの図)

%アルゴリズムの概略図
\begin{figure}[t]
\centering
%  \centerline{\includegraphics[width=8.5cm]{images/algorithm.eps}}
%  \centerline{\includegraphics[width=8.5cm]{images/algorithm3.eps}}
  \centerline{\includegraphics[width=9.5cm]{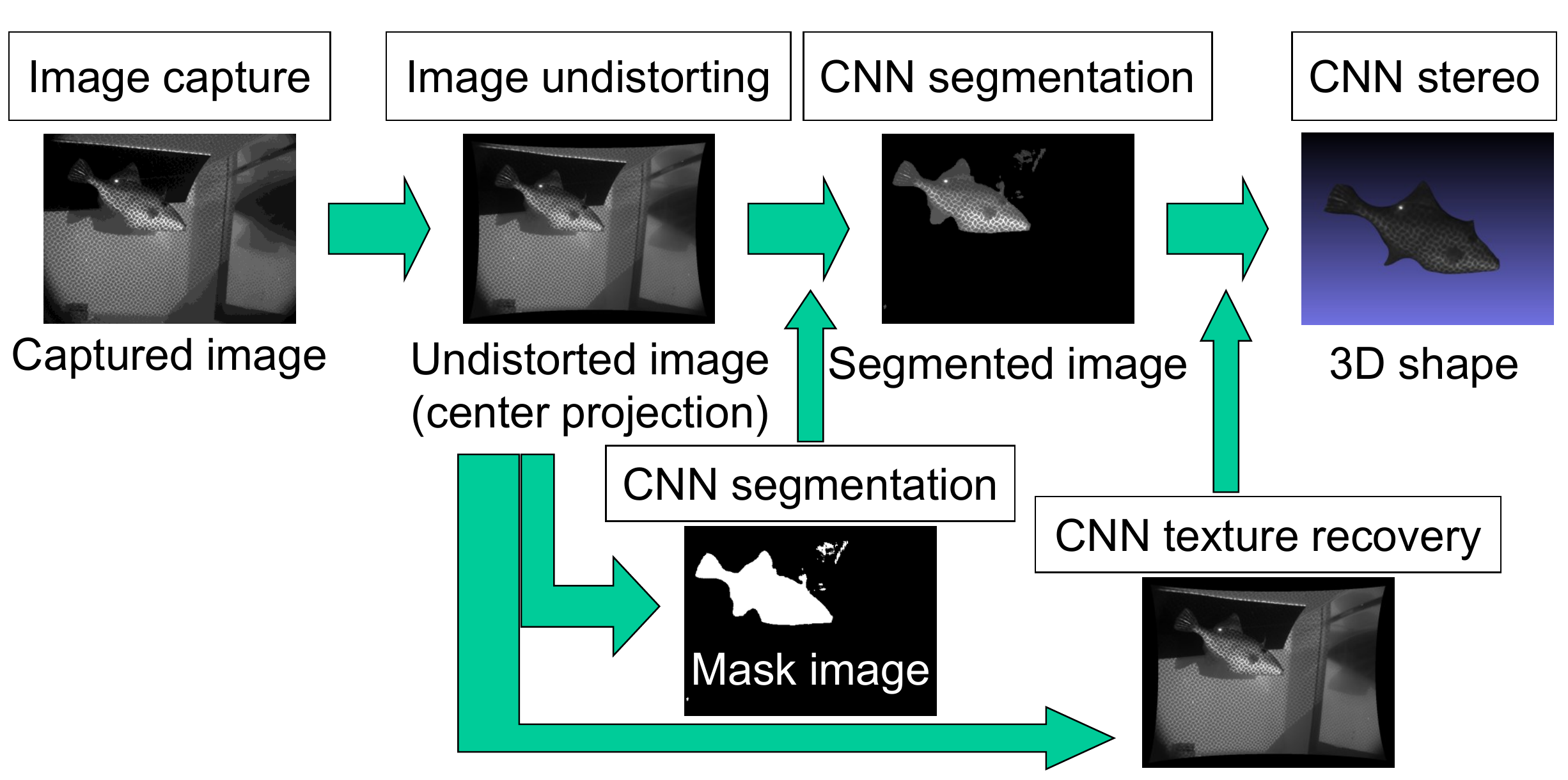}}
%\knoteE{Add texture recovery before 3D shape.}

\caption{Overview of the algorithm.}
\label{fig:overview}
\end{figure}

\begin{figure}[t]
  \centerline{\includegraphics[width=8.5cm]{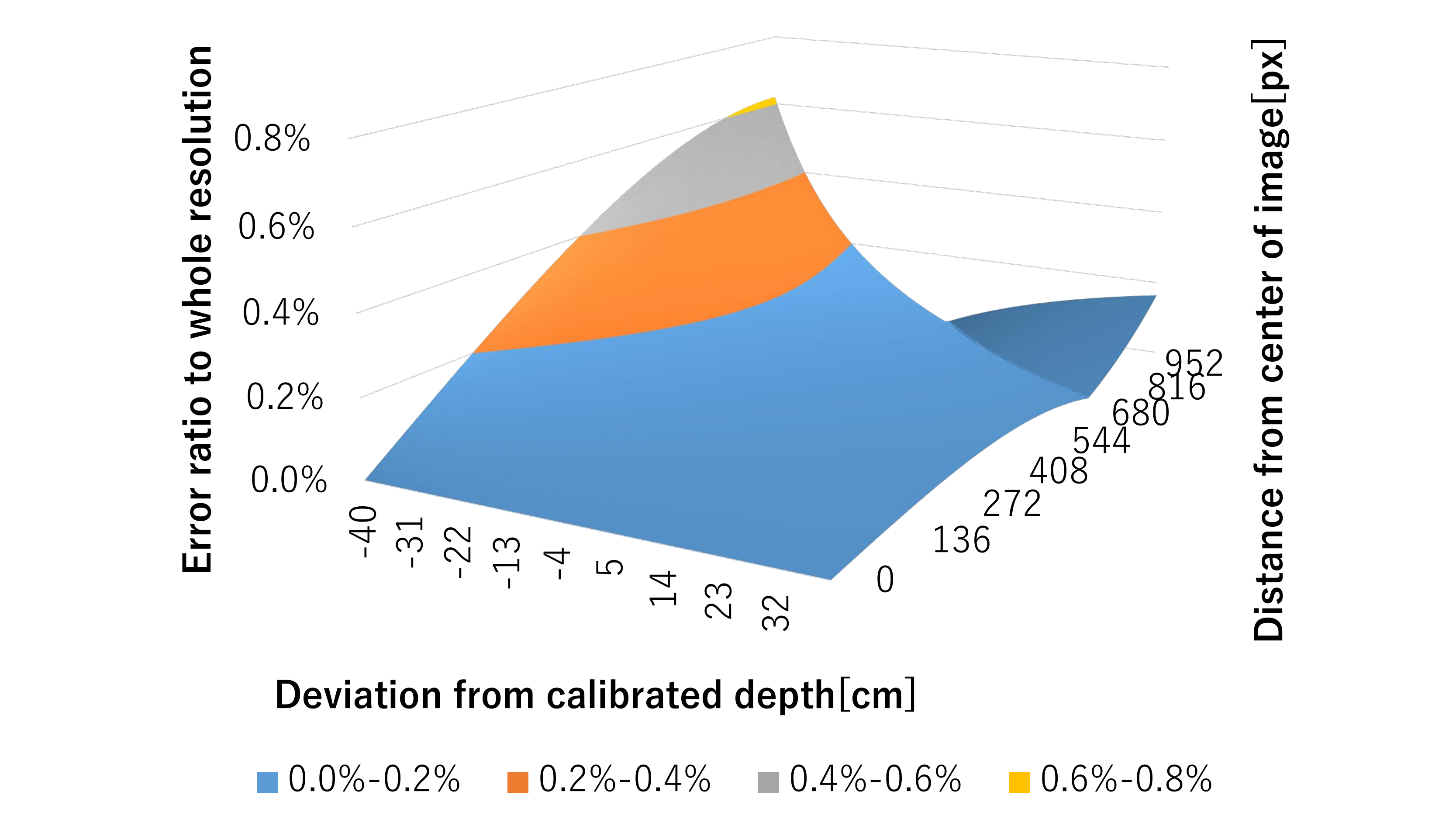}}
\caption{Depth-dependent error of approximation estimated by simulation.}
\label{fig:approx}
\vspace{-0.5cm}
\end{figure}
%(光学モデルの図)

\subsection{Algorithm}
\label{ssec:algo}
The algorithm of our underwater shape reconstruction will be explained by using Fig.~\ref{fig:overview}.
%Now, we describe how our proposal method works(Fig.~\ref{fig:overview}).
First, the camera pair is calibrated. %images for calibration are captured.
The refractions in the captured images are modeled and canceled by 
center projection approximation in our technique using depth-dependent intrinsic and extrinsic 
parameters which are acquired in advance.
In the measurement process, the targets are captured with stereo cameras.
%To capture dynamic scenes, two cameras must be synchronized and capture the scene at the same time.
%A static pattern for stereo matching is projected with projector.
Pattern illumination is projected onto the scene for adding features on it. 
%The captured images are undistorted with the distortion parameters to remove refraction.
%Then, stereo matching is done, but we have one more device.
%Segmentation is applied to rectified images, and mask images are retrieved, which divide objects in images into measurement target and other objects.
From captured images, target regions are detected by a CNN-based segmentation technique, where only fish regions are extracted.
%if we capture fishes. 
%Rectified images are masked with retrieved masks, and get target only images.
Then, a stereo-matching method is applied to the target regions. %segmentation results.
In our technique, a CNN-based stereo is applied to increase stability under the 
condition of dimmed patterns, disturbances by bubbles, and flickering shadows.
%
%done with them, and disparity maps are retrieved for each image couples,
%where correspondences are limited within the detected target regions to increase accuracy.
%Since input images are masked, it becomes harder to take wrong correspondence.
%Masking is also applied to output disparity maps so that unnecessary points will not appear in 3D reconstruction.
%(マスク画像を適用する図)
%
Then, 3D points are reconstructed from the disparity maps estimated by the stereo algorithm.
Outliers are removed from the point cloud and 
meshes are recovered by Poisson equation method~\cite{Kazhdan:EGSGP06}.
Since textures are degraded by bubbles and projected patterns, they are 
efficiently recovered 
by CNN-based bubble canceling and pattern removal techniques.
Using the recovered 3D shapes and textures, 
we can render the dynamic and textured 3D scene. 

%of the captured objects 
%
%are rendered 
%%by 3D CG software 
%using the 3D mesh and the 
%rectified image of the 
%refined texture. % for texture mapping. %ed to the recovered shape.

\subsection{Depth-dependent calibration}
%Projector parameters are not used and they are not necessary.
Because of refractions, captured images of underwater scene are severely distorted.
In this paper, we undistort captured images by a lens distortion model~\cite{Kawasaki:WACV17}.  
%by approximating the effects of refaction 
%
%be approximated by lens 
%
%as if 
%the images become center projection.
%This is favorable for approximation, because it assumes focal point moved backward and there is lens distortion.
%No special consideration is necessary and simply normal calibration is done, and get increased focal distance and lens distortion parameters.
The technique is only an approximation, because refraction effect is not strictly represented as the lens-distortion model, 
but it can be used for stereo matching for limited working distances~\cite{Kawasaki:WACV17}.
% as stated by Kawasaki \etal. 
For the actual process, a calibration tool, \ie, planar board with checker pattern, 
is submerged to a water tank to retrieve intrinsic and extrinsic (camera-to-camera 
transformation) parameters, and thus, it is preferable if the best depth for approximation is 
known in advance.
In the paper, we simulate error using actual parameter of our system as shown in 
Fig.~\ref{fig:approx}, showing that a maximum error is below 
0.8\% if depth range is less than 1m.
%Since refraction approximation depends on depth, approximation error increases as target depth varies from the 
%calibrated depth.
Thus, we set all the devices as close as possible to the water interface 
so that the error becomes small enough to be ignored. 
%Our analysis shows that maximum error is below 
%0.8\% with our system for 1m depth range as shown in Fig.~\ref{fig:approx}.
%According to our prior computation, this pixel gap error falls below 0.1\% of entire resolution at the side of image in normal cases(calibrated depth and target depth is not extremely far away) of our setup, and we prove it through experiment described later.
%After undistortion, images are rectified for stereo matching.

\section{CNN based stereo technique with pattern projection}
\label{sec:cnnstereo}

%To deal with hard cases like bubbles disturbing the images, 
%we introduce several techniques based on CNNs.

%To deal with bubbles and water fluctuation disturbing the images, 
In the technique,
we first apply CNN-based target region extraction technique (\subsecref{segmentation}) to increase robustness as well as decrease calculation time
Then, multi-scale CNN stereo (\subsecref{mscs}) is applied to reconstruct 3D points.

\subsection{CNN-based target-region extraction}
\label{ssec:segmentation}
%The first contribution is CNN-based target-region extraction. 
For many applications, 
reconstruction targets are recognizable, such as swimming
fishes in the water. 
%Limiting the stereo reconstruction processing on the target regions
%improves the stability of the pipeline. 
%
%we use mask image of the target region. 
In general, the wider the range of disparities considered in stereo-matching processes, 
the more ambiguities exist, leading to wrong correspondences.
Thus, by extracting the target regions from the input images and 
reducing possibilities of matching within the detected target regions, 
3D reconstruction process becomes more robust. 
%, since correspondence point of target surface must not be out of target region.

\begin{figure}[t]

\begin{minipage}[b]{0.24\linewidth}
  \centering
  \centerline{\includegraphics[width=2.0cm]{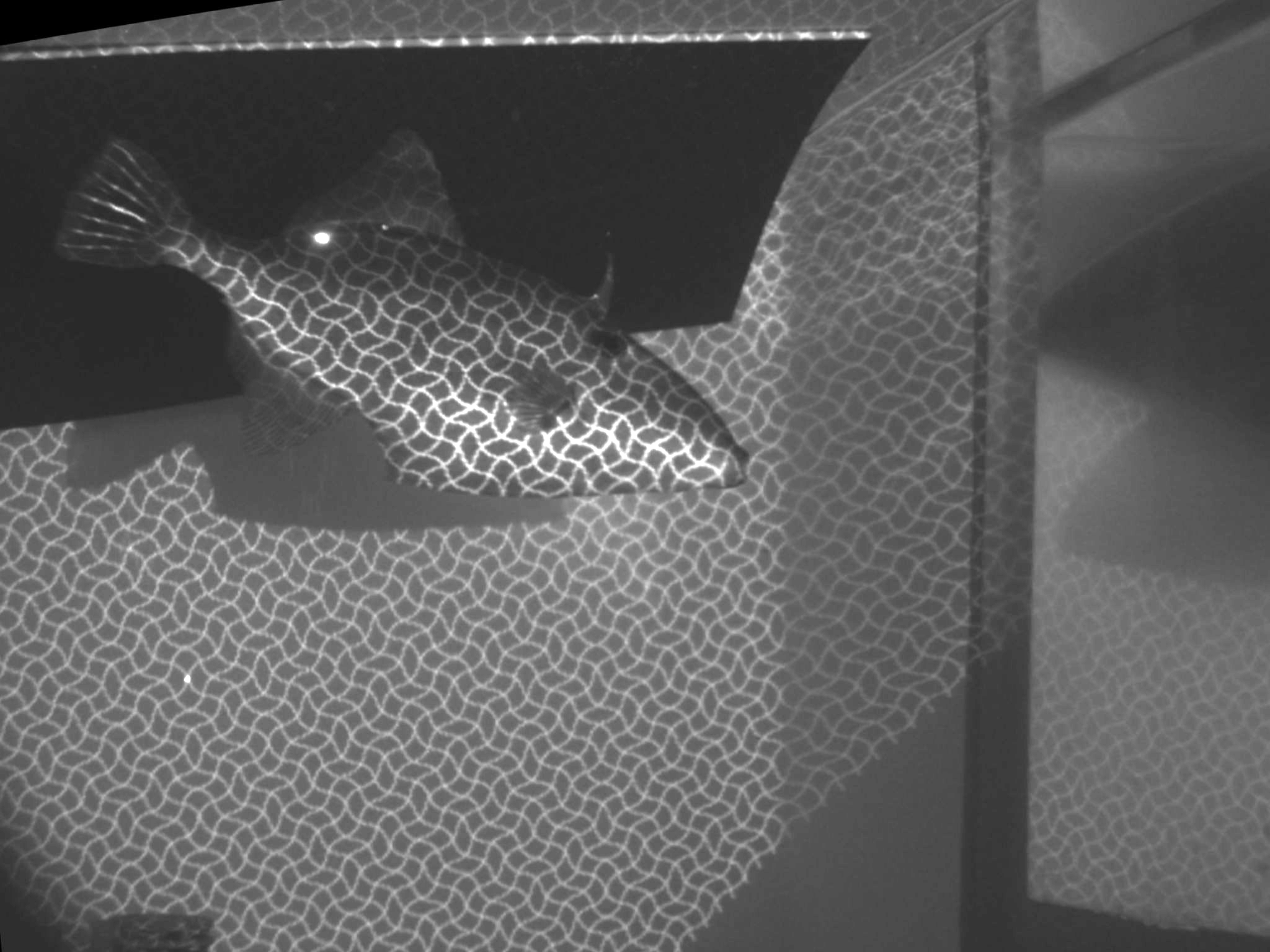}}
\end{minipage}
\hfill
\begin{minipage}[b]{0.24\linewidth}
  \centering
  \centerline{\includegraphics[width=2.0cm]{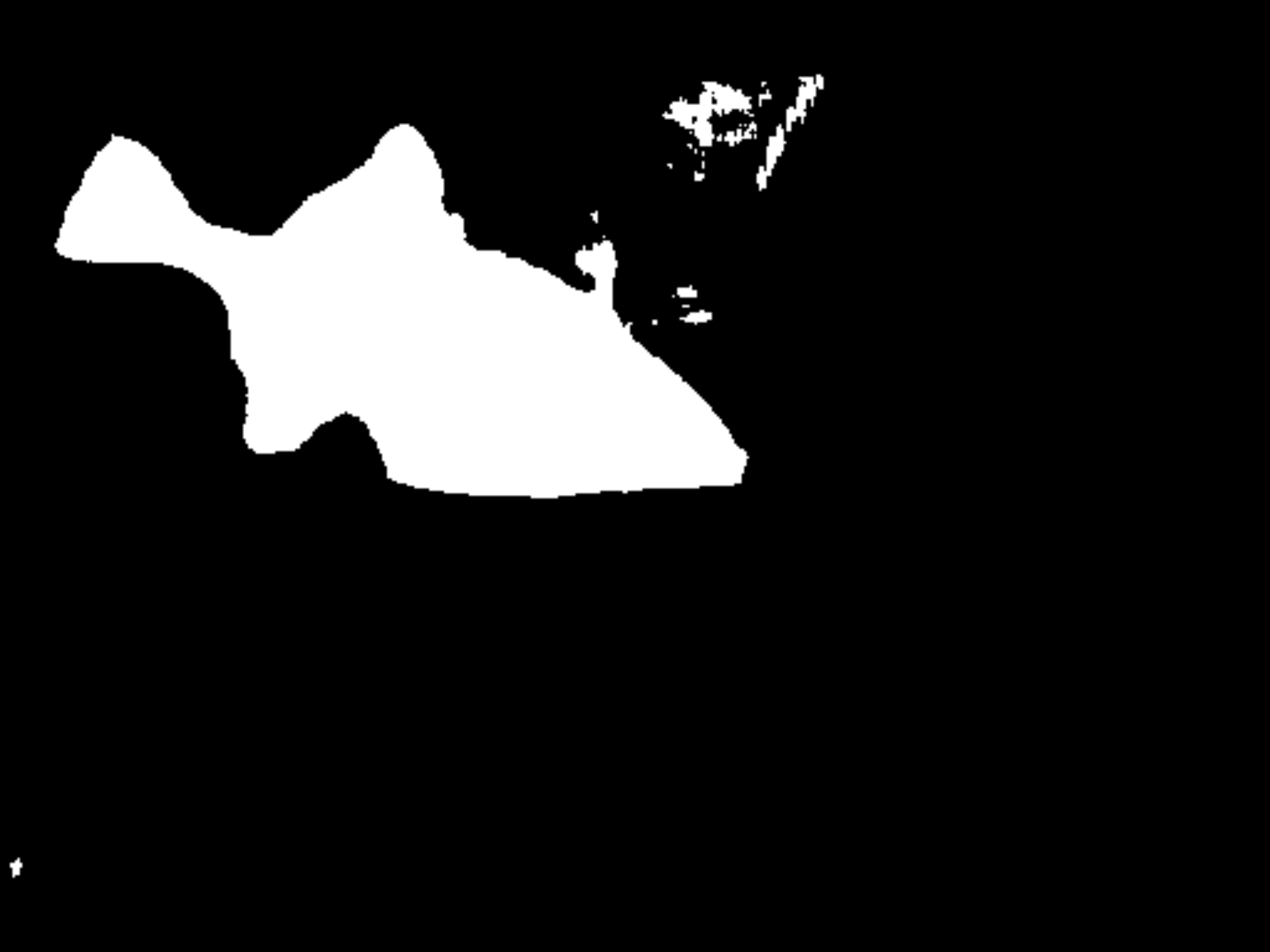}}
\end{minipage}
\hfill
\begin{minipage}[b]{0.24\linewidth}
  \centering
  \centerline{\includegraphics[width=2.0cm]{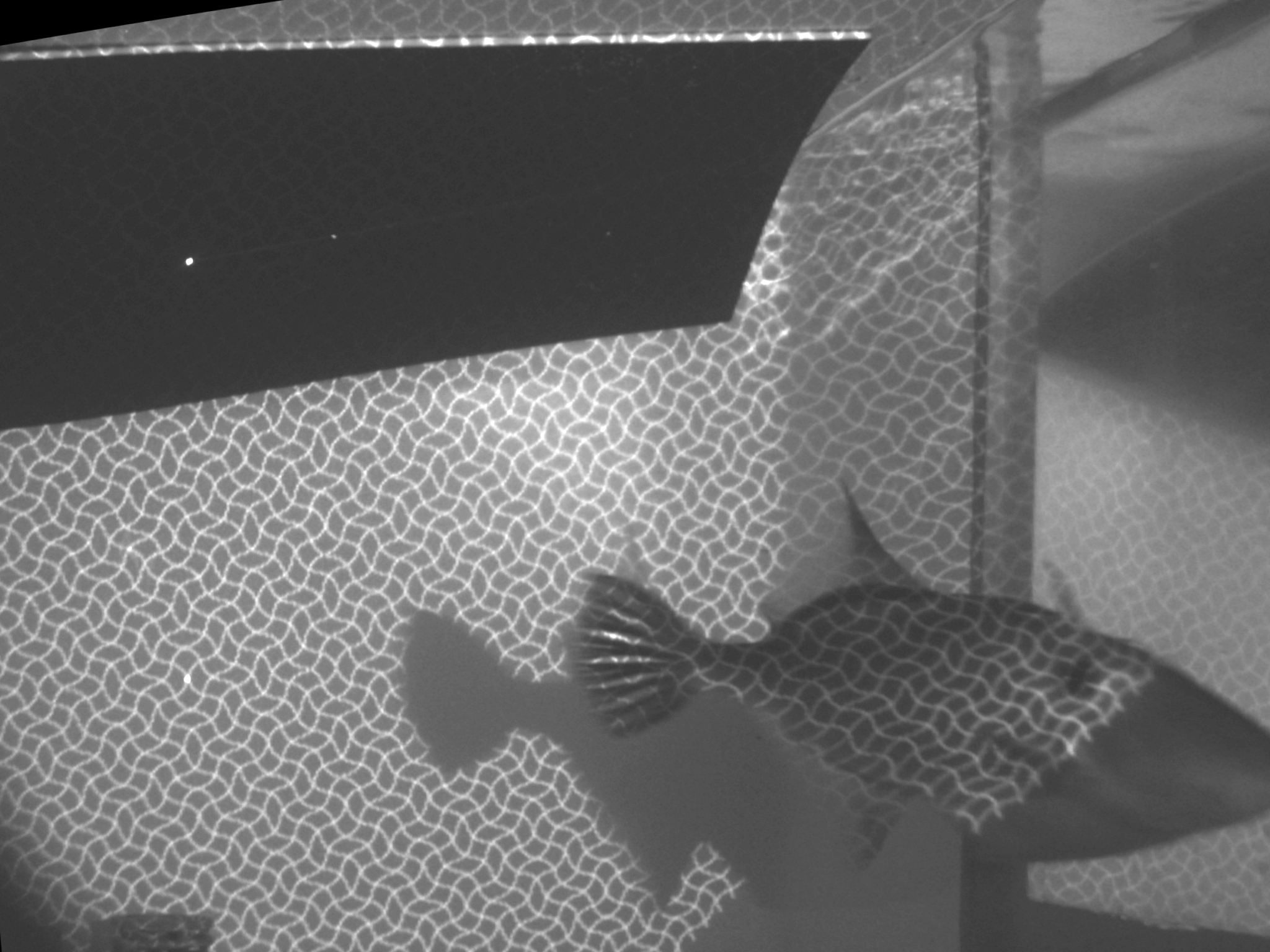}}
\end{minipage}
\hfill
\begin{minipage}[b]{0.24\linewidth}
  \centering
  \centerline{\includegraphics[width=2.0cm]{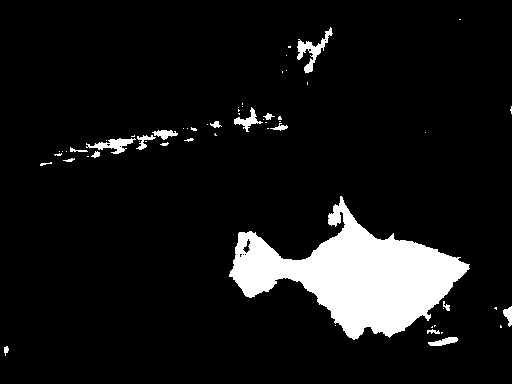}}
\end{minipage}

\caption{ An example of CNN segmentation. {\bf Left:} Successful example. {\bf Right:} Minor failure example. Patternless region was difficult to detect. }
\label{fig:seg}
\vspace{-0.5cm}
\end{figure}

To this purpose, 
%for processing small number of images, we segmented manually, but for large number of images sequentially captured, 
%applied a U-Net
%for this task.
%We segmented objects in captured images into measurement target and other objects.
%For small number of images, we segmented manually, but for large number of images, we had to rely on U-Net.
we implemented an U-Net~\cite{ronneberger2015u}, an FCN with multi-scale feature extraction,
%with Chainer, 
and trained it. % for this task. 
%to capture
%fishes that are illuminated by the pattern projector.
%
We made training dataset from underwater image sequence contains live fish 
(since one of our applications is live fish measurement)
where scenes are illumination by the pattern projector.  
Since both the target and background regions are projected with the same pattern, 
segmentation between those regions was difficult. 
From image sequences, 100 images were sampled and the 
target regions were masked with manual operations.
These training images were augmented by scalings, rotations, and translations.
As a result, we provide 980 pairs of source images and target-region masks for 
training U-Net. 
We used softmax entropy for loss function.
The trained U-Net was tested for large number of images,
 we obtained qualitatively successful results in most examples
 %and in most cases, segmentation results were qualitatively successful 
 (Fig.~\ref{fig:seg}).

In the evaluation process, we have found that the numbers of resolution levels of the U-Net architecture 
is important. 
By using only two or three levels of resolutions, we could not get sufficient results. 
We finally reached the conclusion that the U-Net with five levels of 
resolutions works effectively with our dataset by qualitative evaluation
increasing number of levels. 
%
%About the number of training data pairs, 300 augmented image pairs from around 30 annotated data
%were not enough, such as fishes without pattern illuminations were not detected. 
%The solution becomes improved with increased number of training data, such as 100.  
Regarding the number of training data pairs, 300 augmented image pairs from around 30 annotated data did not work sufficiently for a living fish, but
% without pattern illumination. 
at least 100 pairs were required. % to get a better result. 

Using the obtained results, rectified images are masked so that only measurement target is on the images.
%\knoteE{2 layers: not enough, 5 layers: good output}
%\knoteE{background have same pattern, it makes difficult to segmentation.}
%\knoteE{Learning data: 980pair (original 100 by manual)}
%\knoteE{Fish without texture -- failure -- solution: increase learning data number}
%
%
%
%In this research, we used CNN segmentation and masking on input images, but explicit restriction on stereo matching algorithm is even cleverer implementation.
We also use this mask image to limit the output disparity of stereo matching, 
%within this masked range.
which can drastically decrease calculation time as well as improve accuracy.

\subsection{CNN stereo matching by transfer learning}
\label{ssec:stereo}
In general, normal stereo-matching methods such as SGBM are not robust against 
strong noises since they do not classify pixels into right intensity and wrong intensity~\cite{SGBM}.
%However, since noises are inevitable in real images, 
%they should be coped with. 
%
%it is better to be considered.
Because CNN-based stereo proposed in \cite{mccnn} learns from real images, it is 
possible to cope with the noises.
%We applied the possibility to remove strong noise occurred by bubbles in our method.
\ifx
Fig.~\ref{fig:CNNStereo} is a real example showing CNN-based stereo is robust 
against bubbles than SGBM.
In the method, 
CNNs were used for extracting feature vectors for calculating a similarity measure of 
two image patches. 
\fi
%In the CNN-based stereo, 
In the technique,
small image patches from
stereo image pairs are processed by CNNs and their feature vectors are calculated. 
Similarity measures of the feature vectors are used to find the best-matching 
disparities for every patches of the input images. 

\ifx
In our implementation, 
input images
are limited by the extracted target-regions which are already described.
%Both of them take two images, limited within the extracted target-regions which are already described, and output disparity maps.
We omit post-processing phase of Torch implementation.%, and applied median filters on the output.
Disparity maps are also masked with the extracted target-regions, and clipped by pre-defined disparity ranges.
%The masking processes are also applied for SGBM method for comparison. 
%brightness (Since we know too dark or too bright pixels compared to the neighborhoods are wrong).
%It 
%%helps removing 
%removes wrong disparities caused by bubbles, and makes the disparities a little sparse, but it is better than leaving the errors.
\fi

\begin{figure}[t]
\begin{minipage}[b]{.39\linewidth}
  \centering
  \centerline{\includegraphics[width=3.0cm]{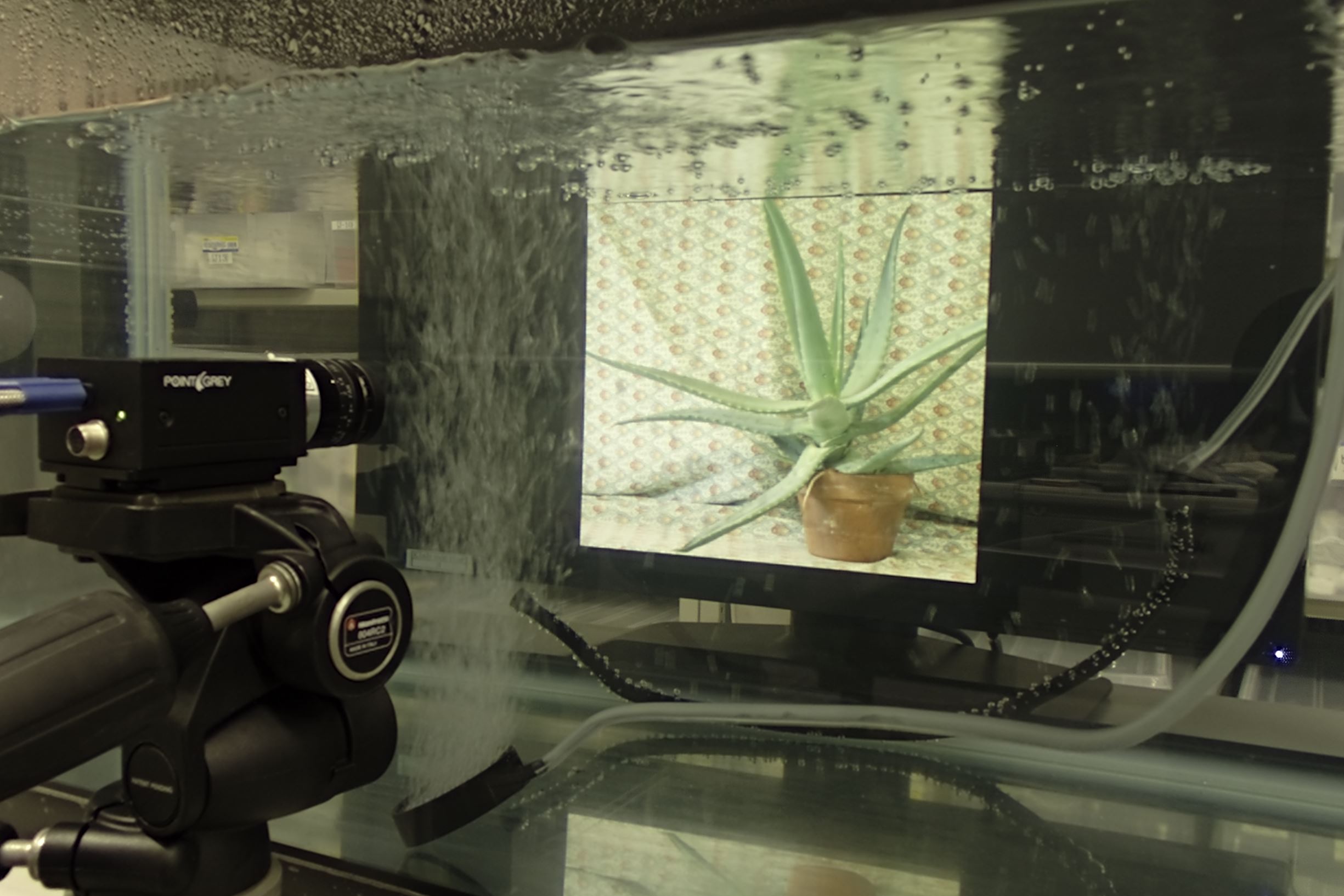}}
(a) Capturing scene
\end{minipage}
\hfill
\begin{minipage}[b]{0.60\linewidth}
  \centering
  \includegraphics[width=2.0cm]{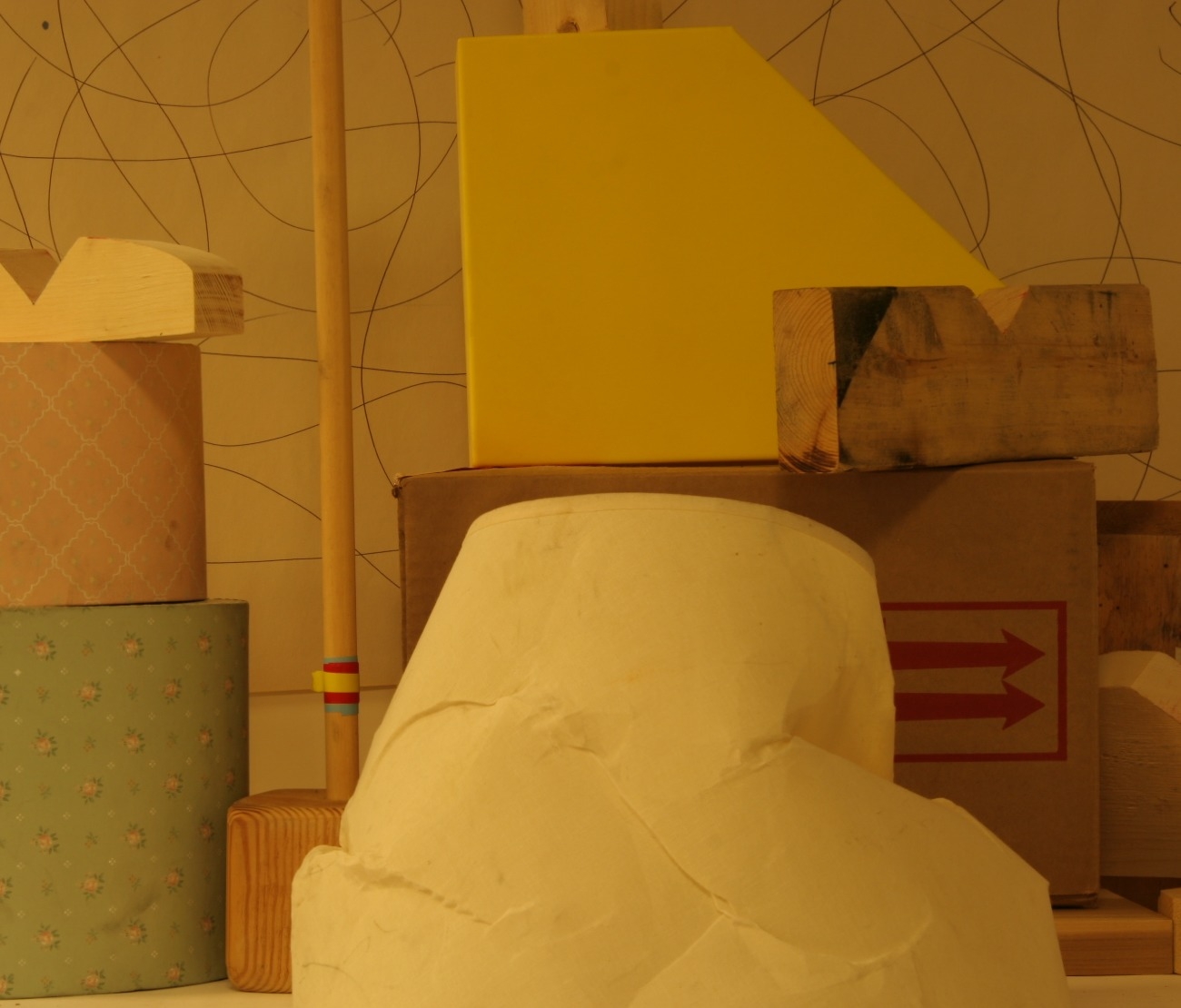}
  \includegraphics[width=2.0cm]{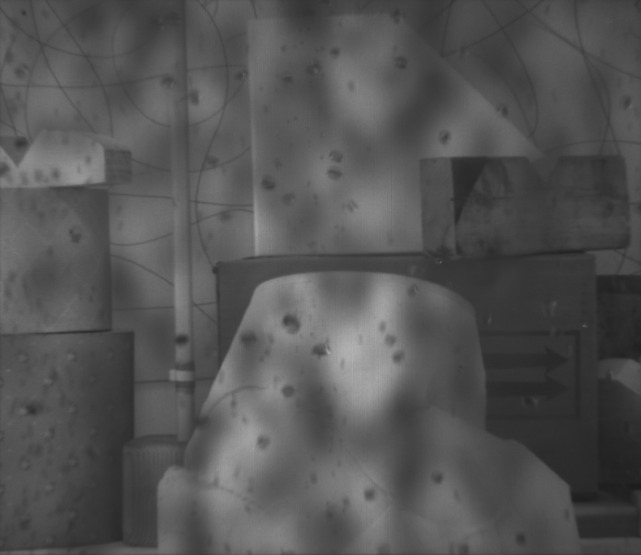}
\\
{(b) Original Middlebury image and with bubbles}
\end{minipage}
%\hfill

%\knoteE{Add example images: configuration figure, without bubble, with bubble1, 2.}

%\knoteE{Right figure moved to Fig.1 or experimental section.}

%\caption{ {\bf Left:} Capturing images through bubbles to create dataset. {\bf Right:} Appearance of our experimental setup from camera view. Projected WAVE pattern can be seen.}
\caption{Capturing images through bubbles to create real learning dataset.}
\label{fig:dataset}
\vspace{-0.5cm}
\end{figure}

In the method, we propose
an effective training method for CNN-based stereo specialized for bubble-disturbed 
images by applying a transfer-learning technique.
%how specialization to bubble-disturbed images 
%is effective through transfer-learning.
%
First, we made a training dataset disturbed by bubbles from Middlebury 2005 and 2006 dataset.
Middlebury dataset contains 1890 images in total, and we used 540 images of them. % (about 33\% of whole training data).
To create images with bubbles, 
we set a display monitor behind
a water tank and put a bubble generator inside a tank (Fig.~\ref{fig:dataset} (left)).
%and a camera
The Middlebury images were presented on the monitor and captured by the camera 
in front of the water tank. % with bubbles.
%Water tank was placed between the display and the camera, air pumps and air stones were used to make bubbles.
%Bubbles appeared in front of the camera, and in front of the display so that both types can be trained. %CNN-based stereo can cope with either types of bubbles.
%
The captured images were warped both by the perspective projection 
and the refraction by the air-water interfaces.
To compensate for this, 
gray code was presented on the display screen and captured by the camera.
Then, lens distortion parameters are estimated, which approximate the refraction, 
by using the gray code.
%the camera was calibrated by using the gray code and 
%the refraction were absorbed by approximation to lens distortion.
%Then, the captured dataset images were undistorted with the lens distortion model,
The captured images were undistorted by the lens distortion parameters
and rectified by homography transformation. %into the original sizes 
Examples of a source image and their bubble-disturbed images are shown in Fig.~\ref{fig:dataset} (right).

Since Middlebury dataset is annotated with ground-truth disparities, we can get positive and negative pairs 
of image patches
for stereo-matching training data.
The positive pairs of patches are sampled from stereo images with corresponding positions, 
whereas the negative pairs are sampled randomly. 
Using these matching pair datasets,  we additionally trained the CNN-based similarity measure pipeline with the captured dataset. %whole dataset containing bubble images.

\subsection{Multi-scale CNN stereo}
\label{ssec:mscs}

%%\knoteII{Rewrite for new network architecture.}

\begin{figure*}[t]
	\begin{minipage}[b]{0.55\linewidth}
		\centering
		\setlength{\fboxsep}{0cm}\fbox{
			\includegraphics[width=9.8cm]{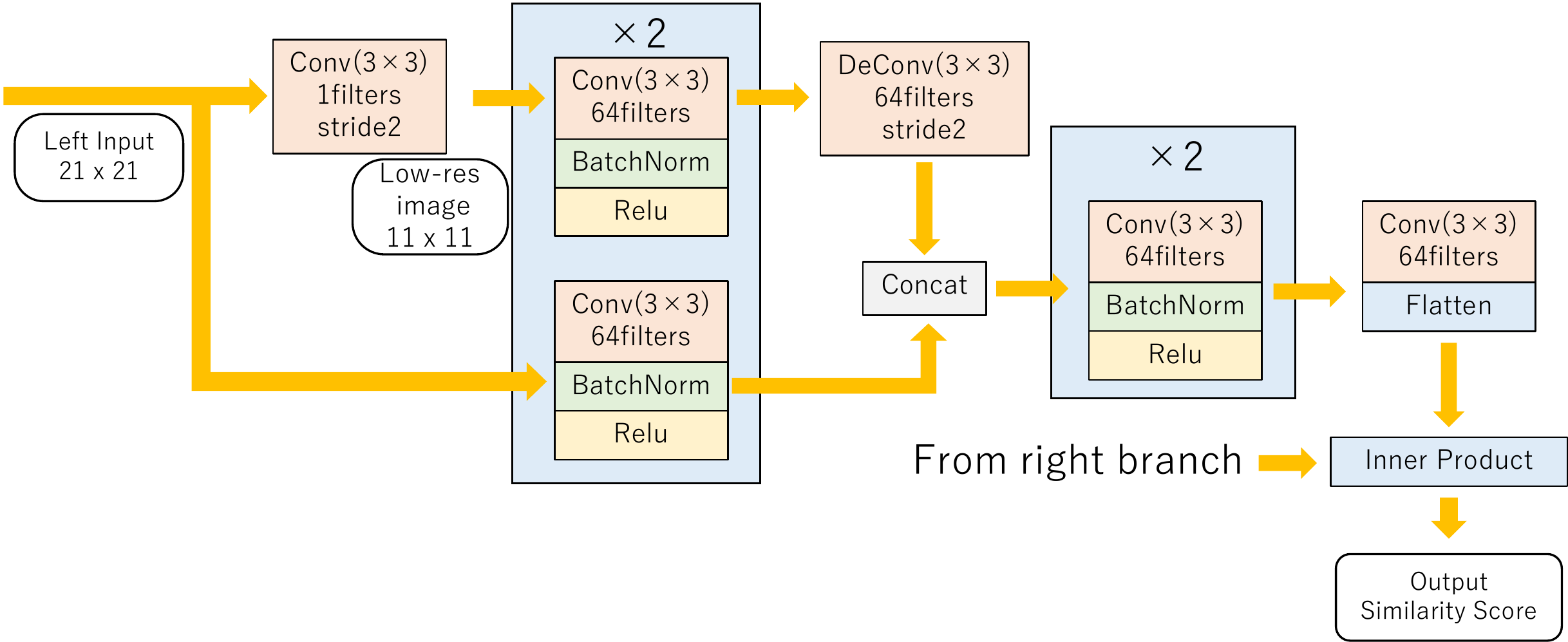}
		}
	\end{minipage}
	\hfill
	\begin{minipage}[b]{0.45\linewidth}
		\centering
		\setlength{\fboxsep}{0cm}\fbox{
			\includegraphics[width=7cm]{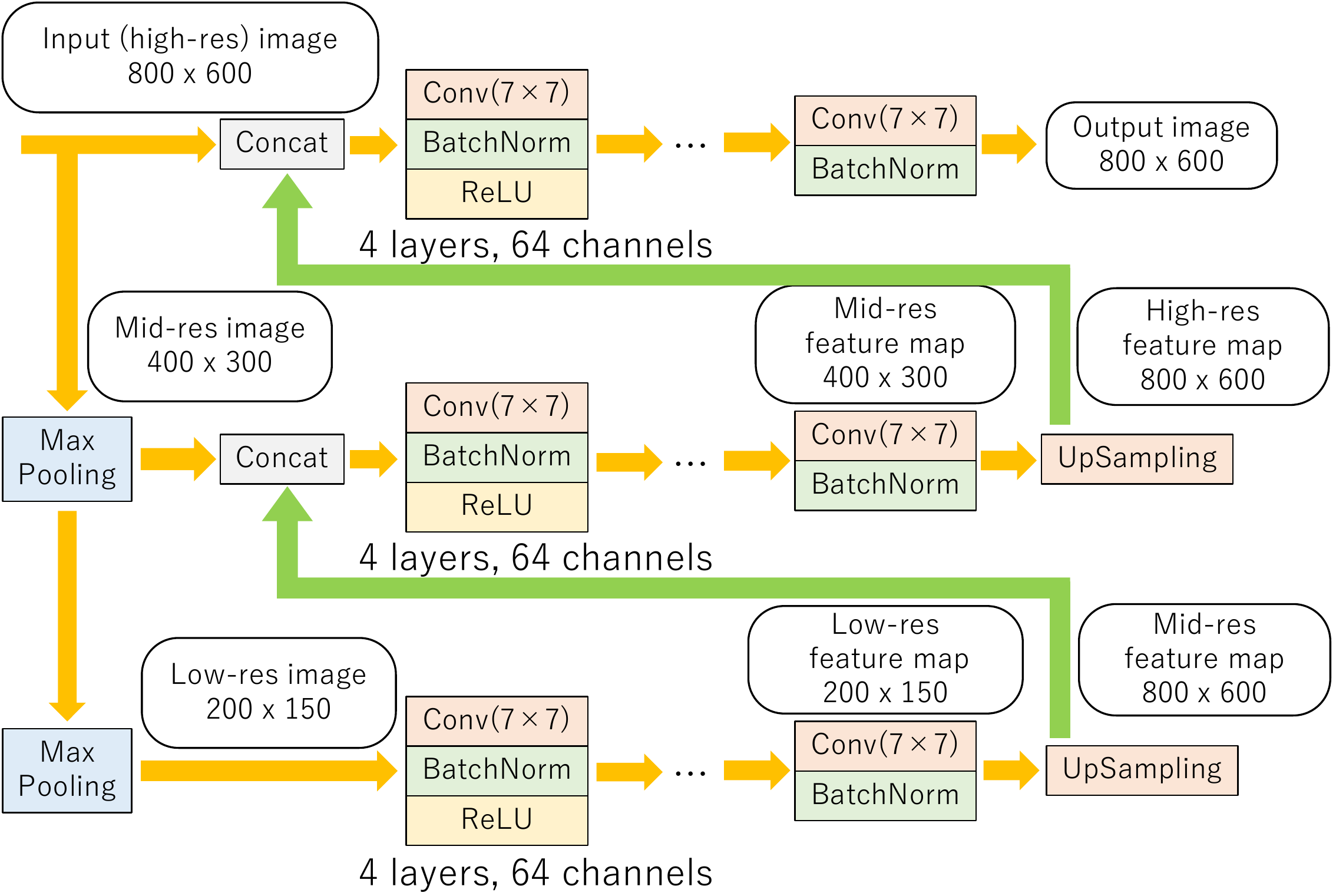}
		}
	\end{minipage}

	\caption{{\bf Left:} Network architecture of multi-scale CNN Stereo. 
{\bf Right:} Network architecture of multi-scale CNN pattern removal. 
Numbers of the data description (round-cornered rectangles) are data dimensions.  }
	\label{fig:network}
\vspace{-0.5cm}
\end{figure*}

CNN-based stereo techniques usually take fixed-size image patches because
%It does not matter in most cases, but sometimes correspondences may be ambiguous without considering wider areas.
a large number of patches with wide variation are trained.
However, it sometimes makes wrong correspondences unless wider regions are 
considered; repetitive pattern of windows are well known example.
Similarly, we assume bubbles whose shapes and sizes vary by large scale, the ambiguity can increase and cause serious failures.
Therefore, we propose a novel network architecture for stereo matching called multi-scale CNN Stereo, which can cope with such ambiguities
(Fig.~\ref{fig:network}(Left)).

The network takes two image patches as input, and outputs similarity score between the patches.
One input patch is processed by two CNN-layer pipelines,
one is for low-resolution, wide-range process, 
and the other is high-resolution, narrow-range process.
The input patch is scaled to half through MaxPooling operation for low-res process, 
and the center sub-image of the input is considered for high-res process.

Each of the convolutional layers is composed of 3$\times$3 convolution, 
batch normalization, and ReLU operation.
%In this implementation, we provide 64 channels of this convolutional layers, 
%where each channels are independent (\ie, the convolutional operations are purely 2D, 3x3).  
As a result, two processed patches (high and low-res) have the same sizes with half the original patches with 64 channels.
The high and low-res results are concatenated, and used as a feature vector to measure similarities. 
%The number of convolutional parameters is 
%$3 \times 3$ $(\mbox{2D convolution})$ $\times 5$ $ (\mbox{4 layers} \mbox{+1})$ $\times 64 (\mbox{channels})$ $\times 2 (\mbox{high and low-res})$ $\times 2 (\mbox{2 input patches})$.
%These parameters 
The neural network parameters
are optimized to  minimize a hinge loss expressed as 
\begin{equation}
	loss = max(0, s_- - s_+ + m),
	\label{eq:loss}
\end{equation}
%expressed in Eq.~\ref{eq:loss}, 
\ie, high similarity score is marked to positive patch pair, while low 
similarity score is marked to negative patch pair, 
where $s_-$ is output score of negative patch pair, $s_+$ is that of positive 
patch pair and
$m$ is margin which means positive score must exceed negative score at this value.
In our training, we used $m = 0.2$ as the margin.

%Processed patches passes separated branches and finally concatenated.
%The branches helps 
Using both high and low-res information helps
recognizing wide area and narrow area similarities at the same time, and it leads to robustness against underwater disturbances.
The ability of Multi-scale CNN Stereo is shown in Fig.~\ref{fig:CNNStereo}.
We trained the multi-scale CNN with training dataset created from modified (\ie, with bubble) Middlebury dataset
similarly with section \ref{ssec:stereo}
with data augmentation of random rotations, scalings, and brightness changes. 
Note that input patches were explicitly extracted from the same epipolar lines of input images in training phase,
but whole image can be inputted in estimation phase.

% the  by extracting patch pairs. 
%Corresponding patch pairs were extracted to make positive samples, and random patch pairs are extracted to make negative samples.
%Extracted patches were randomly rotated, scaled, and changed their contrast and brightness for data augmentation.

%ネットワークの構造図

\section{Texture recovery from noise, bubble and projected pattern}
\label{sec:texture}

For real situations, the captured images are often severely degraded by underwater environments, such as bubble and other 
noises, as well as projected pattern on the object surface.
In order to remove such undesirable effects, we propose a CNN-based texture 
recovery technique.
%denoising and texture recovery technique. 
In our technique, we focus on two major problems, such as bubbles and projected 
patterns. Although those two phenomena are totally different and have different 
optical attributes, it is common in the sense that appearances for both effects have a wide variation in scale.
Note that such wide variation depends on the distance between a target object, bubble and a projector.
Such a large variation of scale makes it difficult for removal by 
simple noise removal method.

Since multi-scale CNN is suitable to learn such a variation, we also use a 
multi-scale CNN for our bubble and pattern removal purpose.
The network for such obstacle removal is shown in Fig.~\ref{fig:network}(Right).
In the figure, it is shown that an original image is converted to three 
different resolutions and trained by independent CNN. Each output is up-sampled 
and concatenated to higher resolution.
%We fixed the input image size as $800\times600$ RGB image, but arbitrary resolution can be handled if trained properly as long as there are sufficient memory.
%Input images are scaled to half size at middle branch, and scaled to quarter size at bottom branch.
%At the end of bottom branch, image is up-sampled to double size and concatenated with the input of middle branch.
%Middle branch is similar network and output image of top branch is the final result.
This network is advantageous because it can handle a large structure of 
projected pattern, as well as it can be trained in a relatively short time.
%At first, each branches are trained to output same images as close to ground truth as possible.
%The network learns important features of projected pattern.
%Then, only top branch is trained to output more accurate result.
%This helps shorten training time.
We prepared two datasets to train the network for bubble removal and pattern removal.
For training bubble removal network, we also used Middlebury dataset containing bubbles mentioned in \subsecref{mscs}. 
For training pattern removal network, we captured several real targets with/without pattern 
projection to create training data. However, the number of data is not 
sufficient to train the network, we synthesize training data by using CG.
We use Middlebury dataset and reconstruct 3D shape with texture map, and then, 
use virtual pattern projector to add pattern onto the object surface.
Then, images were translated, rotated, and scaled randomly for data augmentation.
The pattern removal ability of this network is shown in the experiment.

%実験装置の図
\begin{comment}
\begin{figure}[t]

\begin{minipage}[b]{.48\linewidth}
  \centering
  \centerline{\includegraphics[width=4.0cm]{images/experiment/jpg/IMG_0198.jpg}}
\end{minipage}
\hfill
\begin{minipage}[b]{0.48\linewidth}
  \centering
  \centerline{\includegraphics[width=4.0cm]{images/experiment/jpg/IMG_0200.jpg}}
\end{minipage}

\caption{ Our experimental setup. {\bf Left:} Used water tank. {\bf Right:} Appearance from camera view. Projected WAVE pattern can be seen.}
\label{fig:setup}

\end{figure}
\end{comment}

\section{Experiments}
To evaluate proposed method, we conducted 4 experiments.
In \subsecref{eval}, we describe how our method is accurate and dense under depth-dependent calibration.
In \subsecref{robust}, it is examined that how our multi-scale CNN stereo is robust against underwater disturbances.
In \subsecref{texexp}, qualitative evaluation results of texture recovery are shown.
Finally in \subsecref{demo}, we captured and reconstructed real swimming fish to confirm the feasibility of our method.

%\subsection{Evaluation on accuracy and number of points}
\subsection{Validation of shape reconstruction by depth-dependent calibration}
\label{ssec:eval}
\begin{figure}[t]

  \centering
  \centerline{\includegraphics[width=8.0cm]{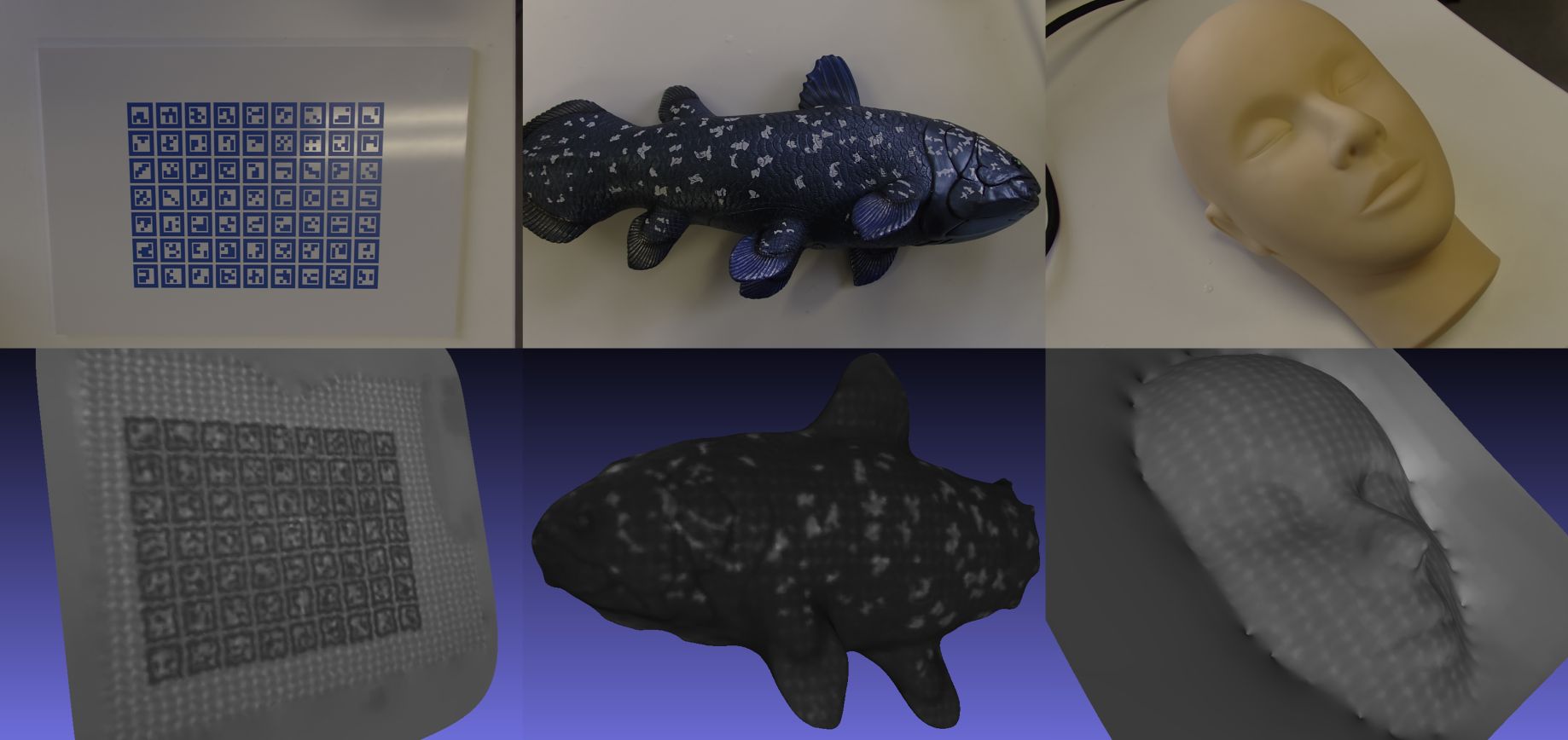}}
  \caption{Upper row shows target objects and bottom row shows reconstruction results. Left to right:
    a calibration board, a vinyl fish and a mannequin head.}
  \label{fig:mesh_reconst}
\vspace{-0.5cm}
\end{figure}

\begin{figure}[t]
  \centering
  \centerline{\includegraphics[width=8.7cm]{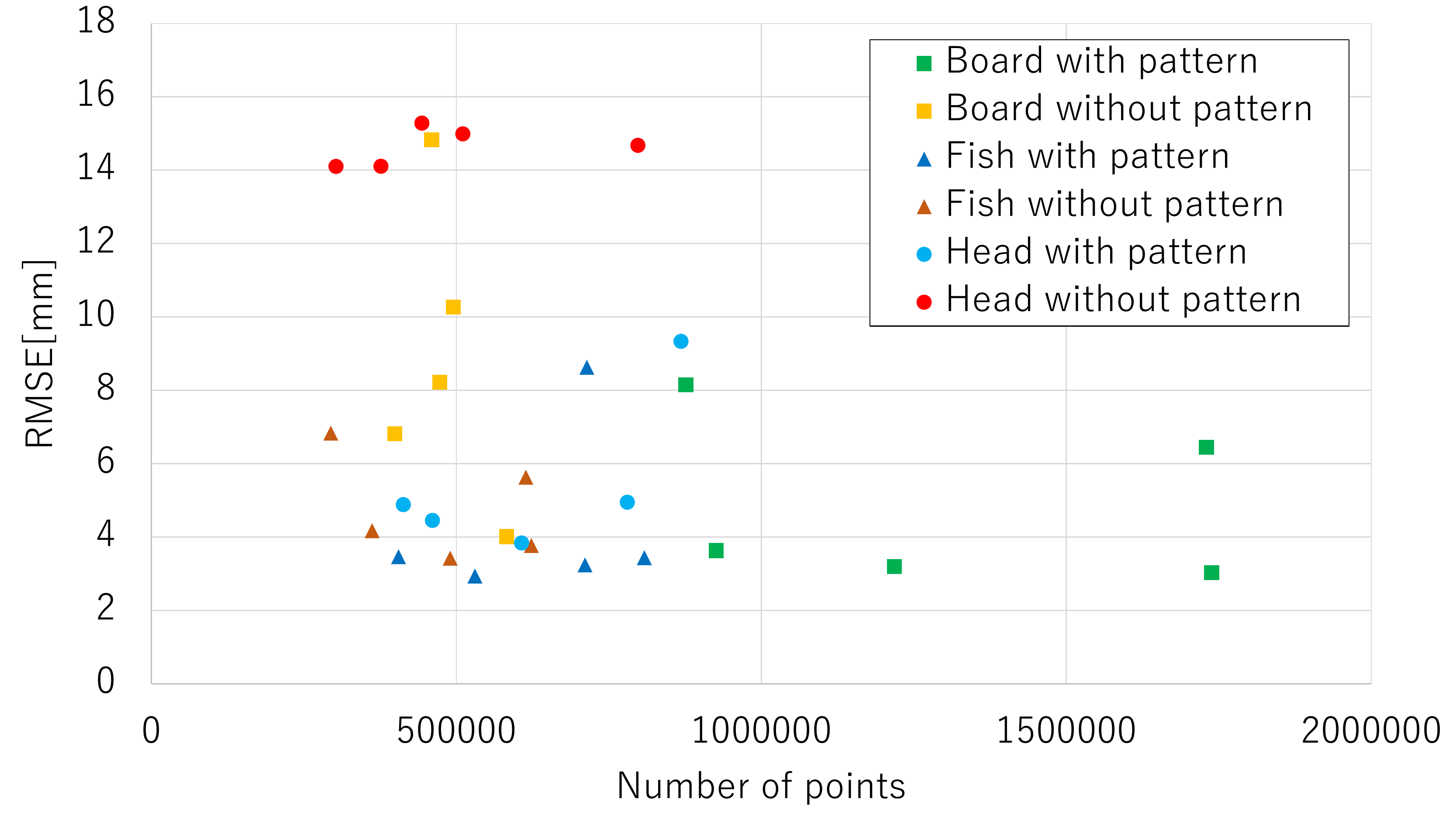}}

\caption{ Graph of accuracy and density experiment. Horizontal axis represents number of reconstructed points, vertical axis represents RMSE from the GT shape, and lower right point is better result. 
Our pattern projection based passive stereo method drastically 
 improves the RMSE as well as point density.}
  \label{fig:eval_result}
\vspace{-0.3cm}
\end{figure}

%We performed experiments to evaluate our method. %(Fig.~\ref{fig:dataset}).
%First, we confirmed accuracy of our method.
%% and effectivity of passive active stereo.
%Second, we tried hard case, bubble scene to examine how CNN-based Stereo is robust under such difficult circumstance.
For the experiments, we used Point Grey Grasshopper3 cameras and Canon LV-HD420 lamp projector.
To reproduce underwater environments, we used a water tank with a size of  90$\times$45$\times$45cm.
Target objects were a calibration board, a vinyl model of fish, and a silicon model of 
a human head as shown in Fig.~\ref{fig:mesh_reconst}. 
They are captured in the air and reconstructed with a structured-light technique to acquire the ground-truth. % using the same camera and projector.
%We captured targets with a structured light technique in air as ground truths using the same camera and projector.
The cameras and the projector were calibrated at a distance of 60cm %with a calibration board 
by our depth-dependent calibration technique and 
captured images were converted to center projection image.
%Calibration RMSE was 0.58 pixel.
Each target object is placed at different  
distances, ranging from 40 to 80cm by 10cm intervals 
and captured with/without pattern projection, \ie, 
% ({\it i.e.} passive stereo), 
in total 180 images were captured.
%Captured images were undistorted, rectified, and segmented manually to remove background.
%Every pair of images were processed by CNN-based stereo, and output disparity maps were smoothed and denoised with median filter(kernel size = 21).
%Disparity maps were also masked again and clipped by proper brightness threshold.
%Finally, we got point clouds from disparity maps.
%Point clouds were cleaned with outlier removal of PCL.
Then, all the objects were reconstructed by the proposed method
%and the output point clouds were cleaned with outlier removal. 
and the numbers of the reconstructed points
% for each mesh after outlier removal 
 and measured ICP residual errors from the ground-truth were calculated.
%Results are averaged values of 3 meshes of each scene.
%For comparison, the same process were applied without projecting the pattern (i.e., pure passive stereo). 
The results are shown in Fig.~\ref{fig:eval_result}.
%Fig.~\ref{fig:mesh} and \ref{fig:graph12}.
It is proved that all shapes are successfully recovered with our depth-dependent calibration technique.
Further, it can be confirmed that, in most cases, a larger number of points were reconstructed with pattern projection than 
without projection.
\ifx
Particularly, textureless objects cannot be reconstructed at all with passive 
stereo, whereas projected pattern made it possible to reconstruct almost entire 
shape. %whole area was reconstructed with the proposed method.
\fi
The accuracies were also better than without 
pattern projection in most cases.
%, especially, stereo matching suffered serious textureless problem and completely failed in head scene without pattern, but accurate matching could be done with pattern.
%It made a major difference between proposal method and passive stereo in ICP error.

%We expected there are little difference in accuracy by distance with proposal method, and we can say approximation error is very small because that was true in most cases.
%In most cases,  the errors caused by refraction approximation were very small.
%However sometimes remarkably big error was confirmed in 40cm scenes,
%where the difference of the real distance and supposed working distance were large. 
%It can be explained 
%wideness of disparity range, since disparity range increases as reciprocal of distance.
%Result of calibration board without pattern in 80cm scene also got large ICP error.
%It is because the board was a little slanted and disparity errors were widely spread.

%精度実験の結果
\begin{figure*}[t]
  \centerline{\includegraphics[width=17cm]{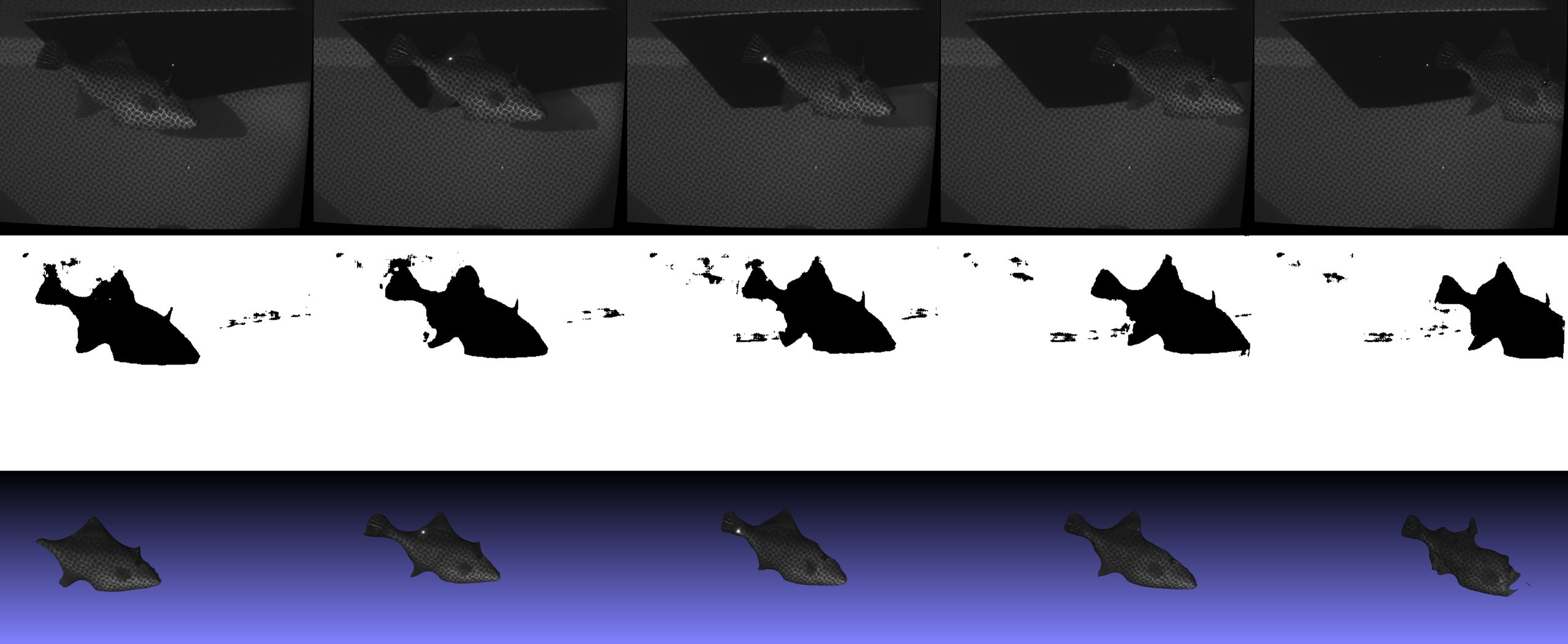}}
%\vspace{-0.4cm}
\caption{Live fish experiment. {\bf Top:} Captured images. {\bf Middle:} Segmentation results. {\bf Bottom:} Reconstruction results.}
\label{fig:demo}

\end{figure*}

\subsection{Evaluation of various CNN stereo techniques}
\label{ssec:robust}
\begin{figure}[t]

  \centering
  \centerline{\includegraphics[width=8.7cm]{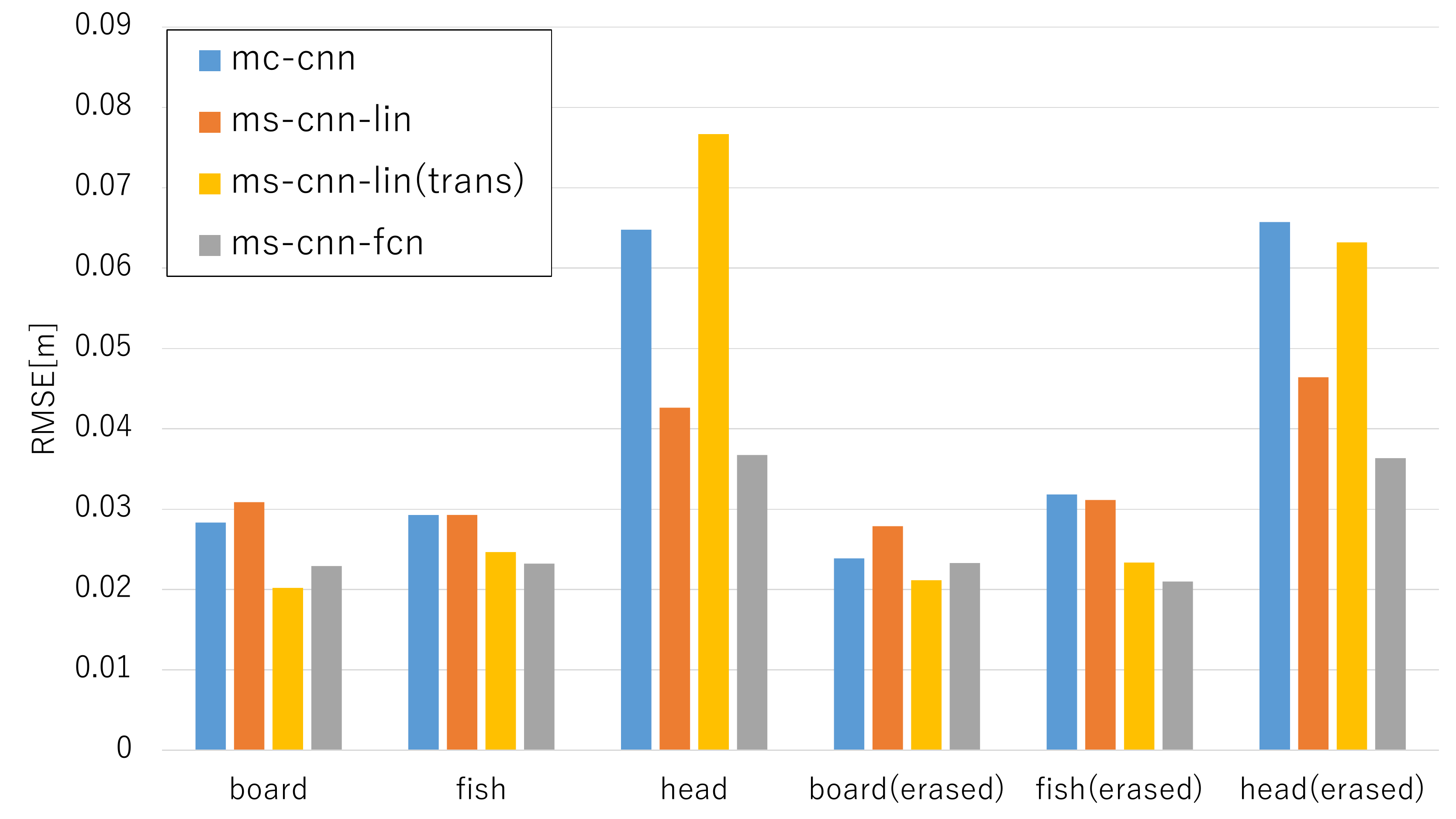}}

\caption{Comparison on proposed method and previous method. Our methods 
    (ms-cnn-fcn) performed best in most cases. 
(erased) means result from bubble removed images. }
%\vspace{-0.3cm}
\label{fig:rmse}
\vspace{-0.5cm}
\end{figure}

\begin{figure*}[h]

  \centering
  \centerline{\includegraphics[width=17cm]{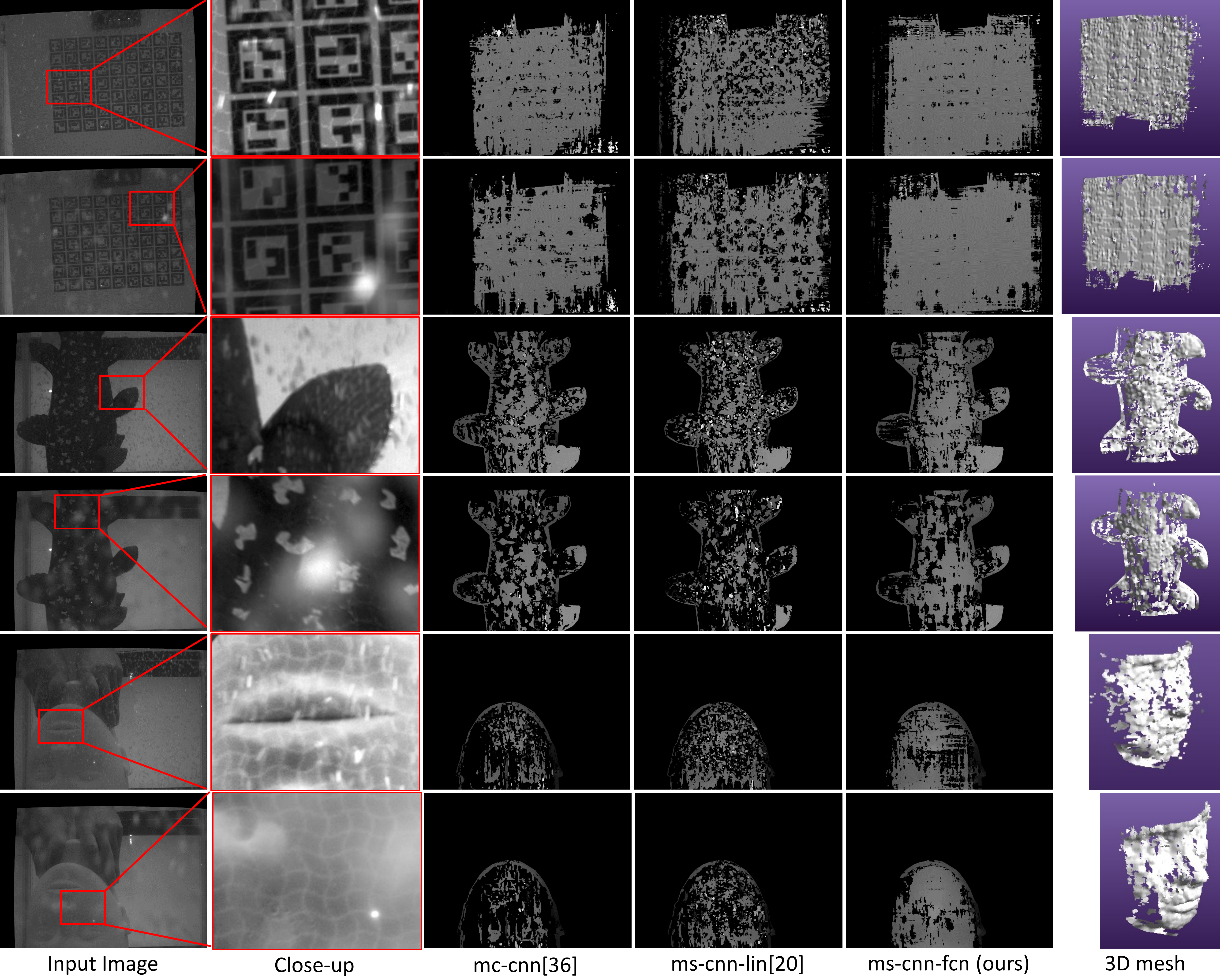}}

%\vspace{-0.3cm}

\caption{ Difference of disparity maps between stereo methods in bubble scene. 
%Robustness of CNN-based stereo against bubble is confirmed.  
Bubble is so severe and almost any method can produce quite poor results, whereas our method
produced much better results.}
\label{fig:CNNStereo}
\end{figure*}

Next, we tested CNN-based stereo for underwater scene with bubbles.
%by processing bubble-disturbed scenes. % to examine how is robust under such difficult circumstance.
%
%Second, we evaluated the CNN-Consecutively we captured bubble scene.
%Two air pumps and air stones for aquarium were placed in the water tank.
%%%%%% moved from section 4.2 %%%%%%
%\knote{moved from section 4.2}
For evaluation purpose, 
%stereo matching, 
we prepared four implementations,
% to compare the results.
such as CNN-based stereo of \cite{mccnn}, 
multi-scale CNN stereo with linear combination (ms-cnn-lin), 
multi-scale CNN stereo with FCN (ms-cnn-fcn),
and transfer learned ms-cnn-lin with bubble erased images (ms-cnn-lin(trans)).
%%%%%% moved from section 4.2 %%%%%%
The target objects were placed at a distance of 50, 60, 70cm and the depth-dependent calibration 
was applied as same as the previous experiment.
We intentionally made bubbles to interfere image capturing process.
We reproduced four bubble environments, \ie, far little bubble, far much bubble, near little bubble, and near much bubble.
In addition, no bubble scenes as reference were prepared.
%tried two positions of the air stones: near cameras and near targets.
%
%We tried two positions of the air stones: near cameras and near targets.
%In near-camera scene, fields of view were completely covered by bubbles, but the 
%scene seemed as if they are entirely blurred since the bubbles are out of focus of the camera. % are transpare.
%In near-target scene, targets were partially covered by bubbles,  and because of their total reflection, we hardly see target surfaces through bubbles.
We captured three pairs of images for each target with five environments.
%position with pattern projection, 
In total, 90 images were captured.
Then, we removed bubbles on the images with multi-scale bubble removal architecture, and reconstructed all the scenes and targets.
We calculated average RMSE from the GT shape of each target.
%Reconstruction process was the same as the above except we used SGBM, transfer-learned CNN and Multi-scale CNN for stereo matching in addition to vanilla CNN-based stereo to compare their robustness.
The results are shown in Fig.~\ref{fig:rmse}.
From the graph, we can confirm that the accuracy of proposed CNN architecture is better than previous method, 
% is slightly better 
%than transfer-learned CNN performed slightly better than the original CNN 
supporting the effectiveness of our method.
Fig.~\ref{fig:CNNStereo} shows examples of the reconstructed 
disparity maps (masked with segmentation results)
 for each technique confirming that shapes are recovered by 
our technique even if captured images are severely degraded by bubbles.

%we counted the number of points and measured the ICP error in a similar way as above as well as registration erro with a threshold,

%In near-camera scene, CNN-based stereo reconstructed almost twice as many points as SGBM, and 
%the accuracies of the results were better than SGBM in most cases.
%We assumed CNN-based stereo is robust against fog-like noise from the results.
%In addition, the transfer-learned CNN performed slightly better than the original CNN 
%supporting the effectiveness of our method.
% density and accuracy in near-camera scenes. %, and that supports our conjecture.

%MSCSの結果を追加

\subsection{Experiments of texture acquisition}
\label{ssec:texexp}
\begin{figure*}[t]

  \centering
  \centerline{\includegraphics[width=17cm]{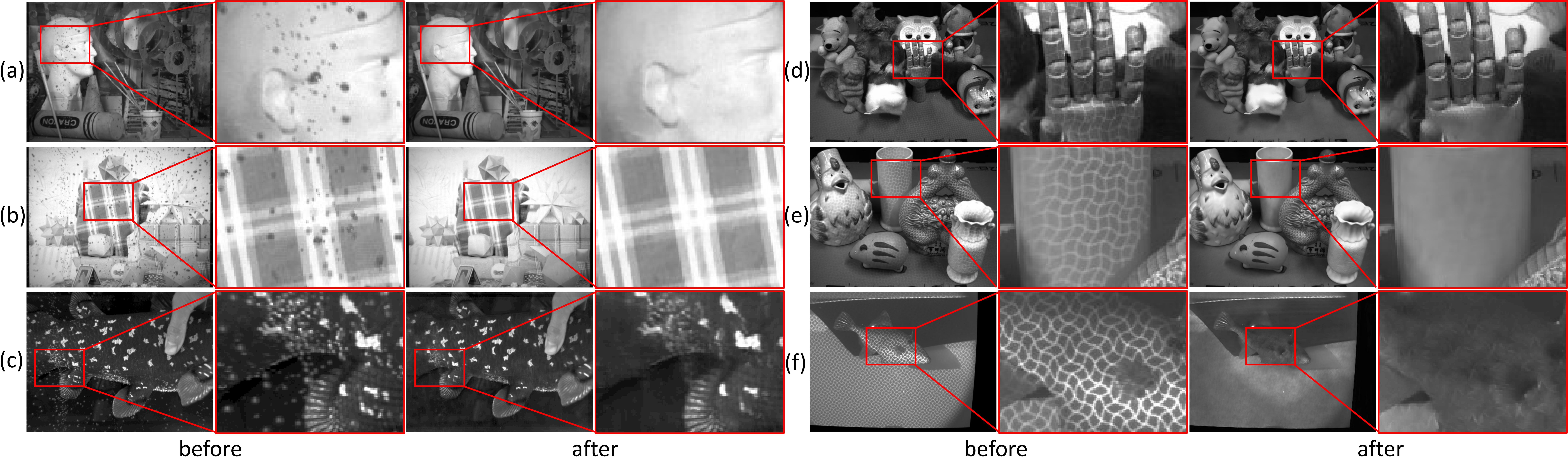}}

\caption{ Result of texture acquisition experiment. Left pane is results of bubble removal and Right pane is results of pattern removal. {\bf Left:} Input images. {\bf Middle-left:} Close-up view of input. {\bf Middle-right:} Output images. {\bf Right:} Close-up view of output. (a, b): Middlebury dataset. (c): Fish model. (d, e): Pattern removal dataset we created. (f): Live fish.}
\label{fig:texexp}

\end{figure*}

We also tested the bubble-removal and the pattern-removal techniques.
%Qualitative evaluation 
The results are shown in Fig.~\ref{fig:texexp}.
%Pattern removal was not applied on underwater images because underwater images we captured were all gray scale.
It is shown that bubbles in the source images were successfully removed as shown 
in the top row of the figure. 
In the bottom row, we can also confirm that projected patterns are robustly  
removed by multi-scale CNN technique.
%al shown 
%the results are reasonably good, although a little traces can be seen in the results. 
%, or the projected light pattern  were effectively removed in the bottom. 

\subsection{Demonstration with a live fish}
\label{ssec:demo}
%Finally, we performed a demonstration experiment.
%Since our priority purpose is to capture dynamic underwater scene, 
Finally, we captured a live swimming fish (filefish) at an aquarium.
We used a special experimental system with an aluminum housing.
Cameras are same as the above experiment, but a projector is substituted by 
laser pattern projector. % with DOE.
We captured and reconstructed 360 frames.
Five frames from the results are shown in Fig.~\ref{fig:demo} for example.
As shown in the figure, we can confirm that the target object is mostly  
successfully segmented by our CNN-based 
object segmentation method. %trained U-Net.
%Randomly selected 100 images from whole data were segmented manually and inflated to make train data, and finally 979 pair of images were used to train the network.
%A little error remained, but it was so small that we can ignore.
In addition, dense shapes of the swimming fish are successfully reconstructed, which proves
the effectiveness and practicality of our method.
Texture are also partially recovered with our method.

\section{Conclusion}
\label{sec:conclusion}

The paper presents a practical underwater dense shape 
reconstruction technique as well as texture refinement method using stereo cameras with a static-pattern projector.
Since underwater environments have severe conditions, such as refraction, light 
attenuation and disturbances by bubbles, we propose a CNN-based solutions, such 
as a target-object 
segmentation, robust stereo matching with a multi-scale CNN and 
CNN based texture-recovery method.
By comparing 3D shape reconstruction with various methods, since other methods 
are severely affected by bubbles and other degradation of underwater environment, 
our method achieved best among them.
Further, bubbles and projected patterns on the objects are successfully removed 
by our method.
We also conducted experiments to show that our approximation of refraction by 
radial distortion is feasible.
%accuracy and robustness of our method.
%Furthermore, we reconstructed live fish and confirmed feasibility and practicability.
Our future plan is to apply the technique to a swimming human for sports analysis.

%\vspace{-0.3cm}
\section*{Acknowledgment}
%%古川科研18A(18H04119)＆川崎科研16B(16H02849)＆川崎科研16国際共同(16KK0151)＆川崎科研18萌芽(18K19824)＆古川カプセル内視鏡MSRA CORE14
%\vspace{-0.4cm}
This work was part supported by grant JSPS/KAKENHI 16H02849, 16KK0151, 18H04119, 
18K19824 in Japan, and MSRA CORE14.
%\vspace{-0.2cm}

\vfill
\pagebreak
\newpage
\clearpage

%\section{COPYRIGHT FORMS}
%\label{sec:copyright}

%You must include your fully completed, signed IEEE copyright release form when
%form when you submit your paper. We {\bf must} have this form before your paper
%can be published in the proceedings.

{\small
\bibliographystyle{ieee}
\bibliography{../bib/shortSTRING,../bib/JabRef,../bib/h-kawa,../180908-eccv-underwater-cnn-active-stereo/refs,201805303dv-r2.bib}
}

\end{document}